\documentclass[a4paper,fleqn]{cas-dc}

\usepackage[numbers]{natbib}
\usepackage{caption}
\usepackage{subcaption}
\usepackage{graphicx}
\usepackage[savepos]{zref}
\usepackage[english]{babel}
\usepackage{multicol}
\usepackage{hyperref}

\definecolor{tableShade}{gray}{1}
\usepackage{multicol} 
\usepackage{multirow}
\definecolor{table_shadow}{rgb} {0.93,0.93,0.93}
\definecolor{table_title}{rgb} {0.8,0.8,0.8}
\usepackage{float}
\usepackage{bm}       
\usepackage{url}
\usepackage{rotating}
\usepackage{siunitx}

\def\tsc#1{\csdef{#1}{\textsc{\lowercase{#1}}\xspace}}
\tsc{WGM}
\tsc{QE}
\tsc{EP}
\tsc{PMS}
\tsc{BEC}
\tsc{DE}

\begin{document}
\let\WriteBookmarks\relax
\def\floatpagepagefraction{1}
\def\textpagefraction{.001}

\shorttitle{LOAM based on Dual Quaternion}

\shortauthors{Velasco et~al.}

\title [mode = title]{DualQuat-LOAM: LiDAR Odometry and Mapping parametrized on Dual Quaternions}
\tnotemark[1]       


\tnotetext[1]{This work has been supported by the grant PID2021-122685OB-I00 funded by MICIU/AEI/10.13039/501100011033 and by ERDF/EU, and grant PRE2019-088069 funded by MICIU/AEI/10.13039/501100011033 and ESF Investing in your future.}

\author[1]{Edison P. Velasco-Sánchez}[orcid=0000-0003-2837-2001]

\fnmark[1]

\ead{edison.velasco@ua.es}

\affiliation[1]{organization={AUROVA, Group of Automation, Robotics and Computer Vision , University of Alicante},
    addressline={San Vicente del Raspeig S/N}, 
    city={Alicante},
    postcode={03690}, 
    country={Spain}}

\affiliation[2]{Worcester Polytechnic Institute, Robotics Engineering,
    addressline={}, 
    city={ Worcester},
    postcode={01609}, 
    country={USA}}

\author[2]{Luis F. Recalde}
\author[2]{Guanrui Li}
\author[1]{Francisco A. Candelas-Herias}
\author[1]{Santiago T. Puente-Mendez} 
\author[1]{Fernando Torres-Medina}

\cortext[cor1]{Corresponding author}

\fntext[fn1]{Edison P. Velasco, Group of Automation, Robotics and Computer Vision (AUROVA), University of Alicante, San Vicente del Raspeig S/N, Alicante, 0369, Spain.}

\begin{abstract}
This paper reports on a novel method for LiDAR odometry estimation, which completely parameterizes the system with dual quaternions. To accomplish this, the features derived from the point cloud, including edges, surfaces, and Stable Triangle Descriptor (STD), along with the optimization problem, are expressed in the dual quaternion set. This approach enables the direct combination of translation and orientation errors via dual quaternion operations, greatly enhancing pose estimation, as demonstrated in comparative experiments against other state-of-the-art methods. Our approach reduced drift error compared to other LiDAR-only-odometry methods, especially in scenarios with sharp curves and aggressive movements with large angular displacement. DualQuat-LOAM is benchmarked against several public datasets. In the KITTI dataset it has a translation and rotation error of 0.79\% and 0.0039°/m, with an average run time of 53 ms.
\end{abstract}

\begin{highlights}
\item A novel LiDAR-only-odometry approach with low drift error based on dual quaternion parameterization of descriptors and optimizer offering a compact and singularity-free representation.

\item Reduced drift error compared to other LiDAR-only-odometry methods, especially in scenarios with sharp curves and aggressive movements with large angular displacement.

\item End-to-end validation of the complete method using the KITTI dataset and our robotic platform comparing the entire approach with others in the state of the art. 

\end{highlights}

\maketitle

\section{Introduction}
In mobile robotics, one of the current challenges is the search for a low-drift and computationally efficient odometry system.  Camera-based systems can encounter difficulties in localization, especially when trying to identify patterns during swift movements, as the resulting images frequently become blurred. Unlike odometry systems based on optical flow from cameras \cite{sun2018robust, alkendi2021state}, LiDAR sensors have proven to be more versatile when working in environments with little or no illumination and sudden changes in brightness. These advantages make LiDAR sensors commonly used for pose estimation in mobile robotics, 
\cite{urmson2008autonomous, levinson2011towards}. Recent technological advances in the field of LiDAR odometry, coupled with decreasing acquisition costs of these sensors, have accelerated research and popularity among the scientific community, making LiDAR odometry one of the preferred options for localization and mapping in complex scenarios \cite{zhang20243d}.

\begin{figure}[ht]
    \centering
    \includegraphics[width=1.0\linewidth]{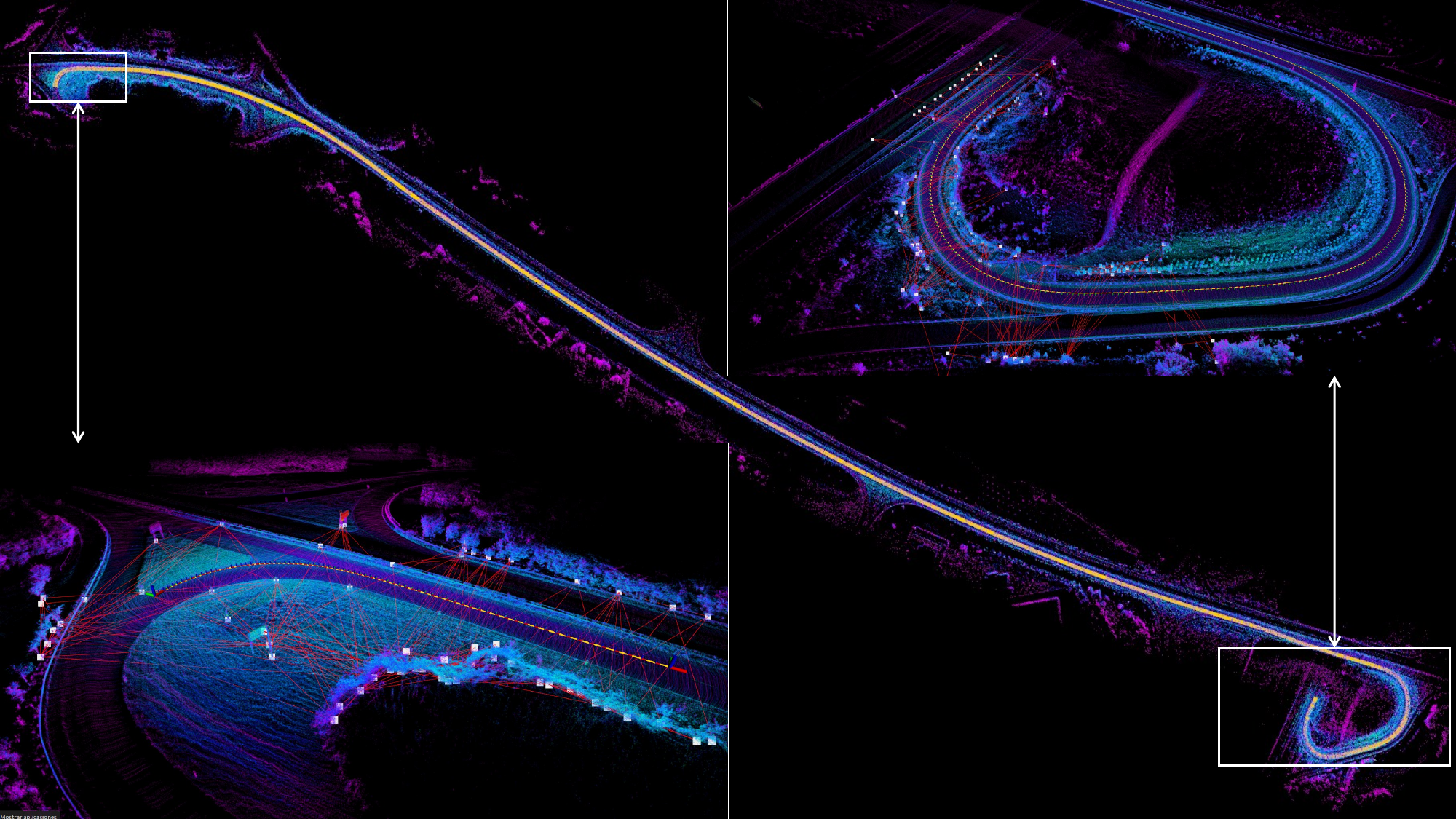}       
    \caption{Results of the proposed DualQuat-LOAM method on Sequence 01 of the KITTI dataset. In the lower left and upper right corner are shown the beginning and end of the sequence respectively, with their corresponding STD map.}  
    \label{fig:dualquat_loam_initial}
\end{figure}

LiDAR-only odometry refers to the technique of determining a robot's pose solely through the analysis of consecutive LiDAR sensor scans \cite{lee2024lidar}. Within this framework, three main approaches for pose estimation are distinguished: \textit{1) Direct Matching}, which directly calculates the transformation between consecutive LiDAR scans. For instance, the ICP (Iterative Closest Point) algorithm \cite{besl1992method} determines the transformation by reducing the error metric between the corresponding points. Nonetheless, this approach can get trapped in local minima, especially in environments with little structure or low overlap between scans. In addition, it can be computationally inefficient due to the large number of iterations required to converge. \textit{2) Feature-Based Matching}, which is based on the extraction of characteristic points from the LiDAR cloud, such as points, edges and planes, to estimate the transformation between LiDAR scans. Thus, the computational load can be reduced by eliminating noise and irrelevant points. However, the performance of this method is highly conditioned by the quality and quantity of the extracted features, which can be challenging in environments with few distinctive features or limited field of view. \textit{3) Deep Learning-Based Matching}, employs deep learning techniques to improve the matching between LiDAR scans. An example of a model employed for LiDAR odometry is LO-Net, which forecasts normal planes and dynamic areas, utilizing neural networks to identify and align key points within LiDAR point clouds. However, these methods require large volumes of training data and are computationally intensive, which may limit their applicability in real time, especially on platforms with restricted computational resources.

Typically, Feature-Based Matching techniques are employed in odometry systems requiring real-time performance and operation on embedded platforms because of their computational efficiency. Nevertheless, these techniques often result in drift errors while estimating the pose over long trajectories. To mitigate this issue, there are supplementary methods that improve these odometry systems, allowing robot location through global feature maps \cite{labbe2013appearance,glocker2014real,data-association,munoz2024geo}. The central aspect of these methods involves matching and minimizing errors when converting between a global descriptor map and the local frame detections aligned to the global frame. For this, it is essential to have a previously established feature map. On the other hand, another approach that reduces drift errors are loop closure algorithms; these allow determining whether the robotic system has passed again through a location previously explored by the LiDAR sensor \cite{chen2021inertial, zhang2021lilo, liu2022light, dellenbach2022ct}. However, the success of loop closure depends on the system's ability to accurately recognize previously visited locations, which can be difficult in dynamic environments or with significant changes in environmental conditions. The types of descriptors used in loop closure can be helpful in LiDAR-only-odometry methods based on Feature-Based Matching. In this context, Stable Triangle Descriptors (STD) \cite{std} are presented as an alternative to LiDAR-based odometry \cite{LTAOM2024}. Unlike features such as edges and surfaces, each STD descriptor has both rotation and translation, making them invariant to rigid transformations, which makes them an option for geometric representation of the environment. Using triangles formed by key points, the STD descriptors capture the local structure and are unchanged in perspective. STD descriptors provide a representation of the environment by incorporating both rotation and translation. However, to effectively minimize pose errors, it is necessary to find an algebraic group that allows compact operation with surfaces, edges and STD descriptors. 

Most LiDAR odometry methods parameterize their operations in the $SE(3)$ group, which allows the representation of rigid transformations.  This group combines rotations and translations in a single mathematical structure, generally represented by homogeneous matrix. Thus, the group is denied as: $SE(3) = \left [  \mathbf{R} ~ \mathbf{t} ~;~ 0 ~ 1 \right ]$, where $\mathbf{R} \in SO(3)$ is rotation matrix and $\mathbf{t} \in \mathbb{R}^3$ is the associated translation vector. This representation employs an overparameterization that generates a high degree of redundancy in relation to the six degrees of freedom (6DoF) \cite{li2020unscented}, thus, redundancies and complexities can be introduced in the optimization problem.  This redundancy causes inefficiencies in memory usage and possible numerical instabilities. On the other hand, a more compact rigid transformation representation that also has no singularities is the use of dual quaternions. This representation has been widely used in the fields of geometry, robotics, and automatic optimization \cite{brodsky1999dual, jia2013dual}, as well as in pose estimation methods \cite{sveier2020dual, uzun2024dual, li2020unscented}. 
    
\begin{figure*}[!ht]
    \centering
    \includegraphics[width=0.9\linewidth]{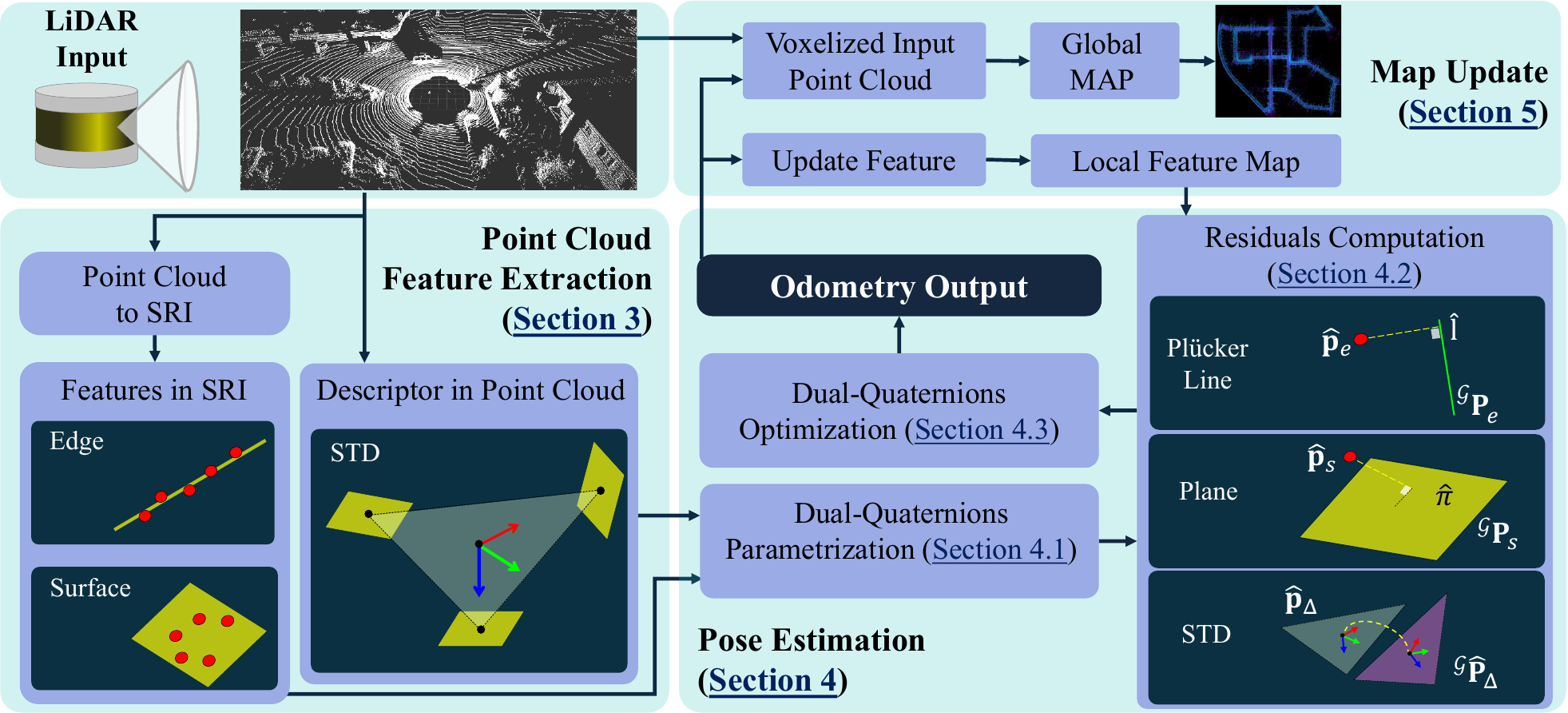}   
        \begin{picture}(0,0)
        \put(-310,108){\hyperlink{pc_feature_extraction}{\phantom{Clickable Area}}}
        \put(-250,05){\hyperlink{Pose estimation}{\phantom{Clickable Area}}}        
        \put(-60,167){\hyperlink{mapping}{\phantom{Clickable Area}}}

        \put(-185,43){\hyperlink{DQ_Parametrization}{\phantom{Clickable Area}}}   
        \put(-190,76){\hyperlink{DQ_optimization}{\phantom{Clickable Area}}}
        \put(-90,120){\hyperlink{residuals}{\phantom{Clickable Area}}}
        
    \end{picture}
    \caption{System overview of our DualQuat-LOAM approach.} \label{fig:pipeline_DualQuat-LOAM}
\end{figure*}
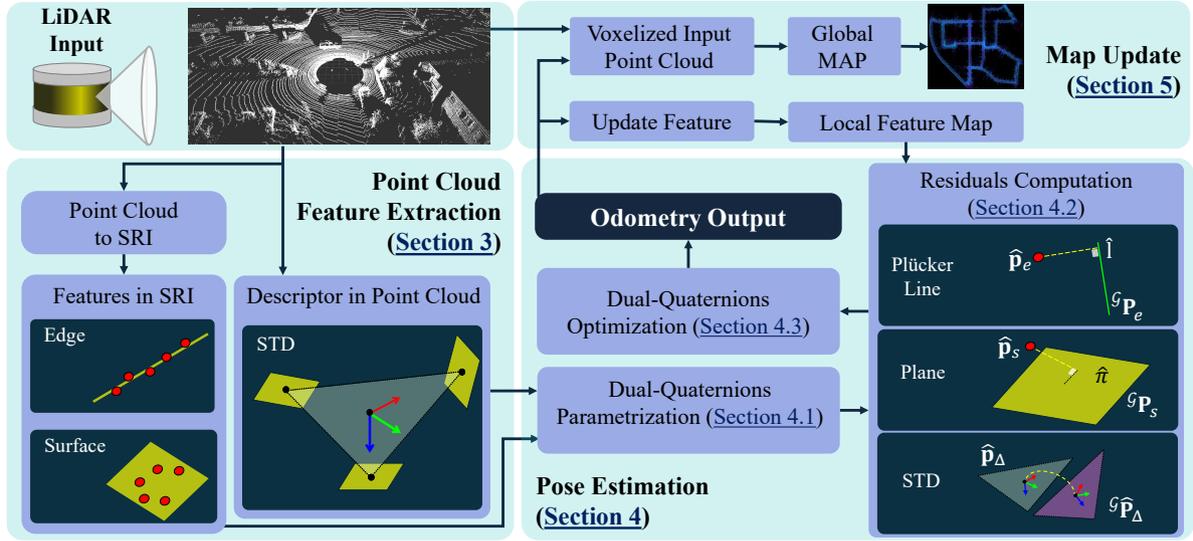

This paper introduces DualQuat-LOAM, a LiDAR-only odometry method that leverages edge, surface, and STD features derived from point clouds to address pose estimation. Incorporating the STD descriptor, which combines both positional and orientational information, enables the integration of pose error into the optimization process for pose estimation. Furthermore, our proposal investigates a parameterized optimization problem for pose estimation using dual quaternions, making it the first LiDAR odometry estimation method to utilize dual quaternion parameterization at the time of writing. Part of our results are shown in Fig. \ref{fig:dualquat_loam_initial}, which were obtained with the sequence 01 of the KITTI dataset, and the complete pipeline of our proposal is shown in Fig. \ref{fig:pipeline_DualQuat-LOAM}.

In summary, our contributions are the following.
\begin{itemize}
\item A novel LiDAR-only-odometry approach with minimal drift error, utilizing dual quaternion parameterization for descriptors and an optimization procedure that provides a compact and singularity-free model.

\item Reduced drift error compared to other LiDAR-only-odometry methods, especially in scenarios with sharp curves and aggressive movements with large angular displacement.

\item End-to-end validation of the complete method using the KITTI dataset and our robotic platform comparing the entire approach with other state-of-the-art methods.   
\end{itemize}

This paper is organized as follows: in Section 2, we present the mathematical preliminaries for the development of the proposed method. In Section 3, we describe the process of feature extraction from the point cloud, where we explain how edge features, surface features and STD descriptors are obtained from LiDAR data. Section 4 focuses on the parameterization of the features within the geometric structure of the dual quaternions, in addition to describing the pose estimation, detailing how the position and orientation of the system are calculated. In Section 5, we explain the map update process, where we describe how new point cloud information is integrated into the global map, as well as into the edge feature, surface, and STD descriptor maps. Section 6 presents the experiments and results obtained, showing comparisons with other state-of-the-art methods in various experimental scenarios, both with data from public datasets and with data generated by our development platform. Finally, Section 7 presents the conclusions of this work and suggests possible lines of future research.

\section{Mathematical Preliminaries}

This section presents an introduction to the concepts and operations of dual quaternions. This information is based on the concepts presented in \cite{adorno2020dq, duales_quat}.

\subsection{Quaternions}
The unit quaternions are an extension of  complex numbers with a compact set defined as:
$$
 \mathbb{H} \triangleq \left \{ q_1 + {\mathbf{i}} q_2  + {\mathbf{j}} q_3  + {\mathbf{k}} q_4 : q_1, q_2, q_3, q_4 \in \mathbb{R} \right \} 
$$
where the imaginary units ${i}$, ${j}$, and ${k}$ satisfy $\mathbf{i}^2 = \mathbf{j}^2 = \mathbf{k}^2 = \mathbf{i}\mathbf{j}\mathbf{k} = -1$. We represent a quaternion using the mathematical accent $(\tilde{\;\;})$ combined with bold font. Consequently, a quaternion can be expressed in the following manner $\tilde{\mathbf{q}}\in  {\mathbb{H}}~;~\tilde{\mathbf{q}} = q_1 + {\mathbf{i}} q_2  + {\mathbf{j}} q_3  + {\mathbf{k}} q_4~=~ \text{Re}(\tilde{\mathbf{q}}) + \text{Im}(\tilde{\mathbf{q}})  $. The component $\text{Re}(\tilde{\mathbf{q}})$ denotes the real part, while $\text{Im}(\tilde{\mathbf{q}})$ corresponds to the vector comprising the imaginary part of the quaternion. The quaternion conjugate is formulated as $\tilde{\mathbf{q}}^*$, achieved by reversing the signs of the imaginary elements while maintaining the scalar component unchanged as follows $\tilde{\mathbf{q}}^{*} = \text{Re}(\tilde{\mathbf{q}}) - \text{Im}(\tilde{\mathbf{q}})$.

Additionally, a vector $\mathbf{a} \in \mathbb{R}^3$ can be expressed as an element of a subset of quaternions that only contains the imaginary part, referred to as pure quaternions; this subset can be defined as: \mbox{$\mathbb{H}_p \triangleq \{ \tilde{\mathbf{a}} \in \mathbb{H} : \textrm{Re}(\tilde{\mathbf{a}}) = 0 \}$}.
Quaternion multiplication is defined as $\otimes$, taking into account the properties of imaginary units; subsequently, the cross product $\times$ is defined as:
$$
\tilde{\mathbf{q}}_1 \times \tilde{\mathbf{q}}_2 = \frac{\tilde{\mathbf{q}}_1 \otimes \tilde{\mathbf{q}}_2 - \tilde{\mathbf{q}}_2\otimes \tilde{\mathbf{q}}_1}{2}
$$
where $\tilde{\mathbf{q}}_1, \tilde{\mathbf{q}}_2 \in \mathbb{H}_p$.

The quaternions can represent rotations of a rigid body which are called unitary quaternions and are defined as:

$$
\mathbb{S}^{3} \triangleq \{ \tilde{\mathbf{q}} \in \mathbb{H} : \| \tilde{\mathbf{q}} \| = 1\}
$$

\subsection{Dual Quaternions}
Dual numbers  are an extension of real numbers, commonly used in the fields of geometry, robotics, and optimization \cite{brodsky1999dual, jia2013dual}. A dual number $\bar{ \mathbf{d}} \in \mathbb{D}$ is described as $\bar{\mathbf{d}} = \mathbf{a} + \epsilon \mathbf{b}$, where $\mathbf{a}$ and $\mathbf{b}$ are elements that denote the primary and dual components of $\bar{ \mathbf{d}}$, respectively, and $\epsilon$ is the dual operator that conforms to $\epsilon^2 = 0$ and $\epsilon \neq 0$. Generally, the primary and dual components are composed of similar types of elements, such as scalars, complex numbers, or quaternions.

We represented a dual quaternion using the mathematical
accent $(\hat{\;\;})$ followed by a bold font. A dual quaternion $\hat{\mathbf{q}} \in  {\mathcal{H}}$ is an extension of dual numbers and quaternions. The dual quaternion set can be defined as:
$$
\mathcal{H} \triangleq \left \{ \tilde{\mathbf{q}}_\mathbf{p} + \epsilon \tilde{\mathbf{q}}_\mathbf{d} : ( \tilde{\mathbf{q}}_\mathbf{p}, \tilde{\mathbf{q}}_\mathbf{d}) \in \mathbb{H}, \epsilon^2 = 0, \epsilon \neq 0 \right \}
$$
Dual quaternions can also be formulated as: $\hat{\mathbf{q}} = \tilde{\mathbf{q}}_\mathbf{p} + \epsilon \tilde{\mathbf{q}}_\mathbf{d}$;  where $\tilde{\mathbf{q}}_\mathbf{p} = \mathcal{P}(\hat{\mathbf{q}}) $ is called the primary part  and $\tilde{\mathbf{q}}_\mathbf{d} = \mathcal{D}(\hat{\mathbf{q}})$ represents its dual part.  
Multiplication of two dual quaternions $\hat{\mathbf{q}}_1 ,\hat{\mathbf{q}}_2$ is represented by $\boxtimes$, and defined as:  $\hat{\mathbf{q}}_1 \boxtimes \hat{\mathbf{q}}_2 = \tilde{\mathbf{q}}_1 \otimes \tilde{\mathbf{q}}_2 + \epsilon \left ( \tilde{\mathbf{q}}_1 \otimes \tilde{\mathbf{q}}_2 +  \tilde{\mathbf{q}}_2 \otimes \tilde{\mathbf{q}}_1\right )$.

Moreover, there are three types of conjugates for dual quaternions \cite{SE3kalmanfilter}:
\begin{itemize}
    \item Dual conjugate: $\hat{\mathbf{q}}^{1*} = \tilde{\mathbf{q}}_\mathbf{p} - \epsilon\tilde{\mathbf{q}}_\mathbf{d}$
    
    \item Full conjugate: $\hat{\mathbf{q}}^{2*} = \tilde{\mathbf{q}}_\mathbf{p}^* + \epsilon\tilde{\mathbf{q}}_\mathbf{d}^{*}$
    
    \item Real conjugate: $\hat{\mathbf{q}}^{3*} = \tilde{\mathbf{q}}_\mathbf{p}^{*} - \epsilon\tilde{\mathbf{q}}_\mathbf{d}^{*}$
\end{itemize}

Same as a pure quaternion, the set of pure dual quaternions can be defined as: $\mathcal{H}_p \triangleq \{ \hat{\mathbf{q}} \in \mathcal{H}: \textrm{Re}(\hat{\mathbf{q}}) = 0 \}$. Furthermore, a vector $\mathbf{a} \in \mathbb{R}^3$ can be represented as a pure dual quaternion as follows: $ \hat{\mathbf{a}} \in \mathcal{H}_p~;~\hat{\mathbf{a}} = \mathbf{0} + \epsilon\tilde{\mathbf{a}}~;~\tilde{\mathbf{a}} \in \mathbb{H}_p$.

The dual quaternion exhibit the properties of both imaginary and dual units, and one of their main uses is the compact and efficient representation of rigid body movements \cite{adorno_dualquat},  which are denoted as unitary dual quaternions and defined as:

$$
\mathcal{S} \triangleq \{ \hat{\mathbf{q}} \in \mathcal{H}:  \| \hat{\mathbf{q}} \| = 1\}
$$
Where $\hat{\mathbf{q}} \in \mathcal{S}$ is a unit dual quaternions. It can be expressed as a translation $\mathbf{t} \in \mathbb{H}_p$ and  a rotation $\mathbf{r} \in \mathbb{S}^{3}$, yielding the following representation:
$$
\hat{\mathbf{q}} = \mathbf{r} + \epsilon \frac{1}{2} \mathbf{t} \otimes \mathbf{r}
$$

Additionally, we introduce the following definitions as:
vectors are represented by bold quantities as: $\mathbf{p} \in \mathbb{R}^n$;  matrices are indicated in bold capital letters as $\mathbf{P} \in \mathbb{R}^{n\times m}$. Finally, to identify an element within a reference frame, we employ a prefix attached to the element such that $^\mathcal{L}\mathbf{p}$; the prefix $\mathcal{L}$ being the frame of reference of the vector $\mathbf{p}$. Throughout the article, we establish the following definitions: an global frame denoted as $\mathcal{G} = \lbrace \bm{c}_\mathcal{G}, \mathbf{g}_x, \mathbf{g}_y, \mathbf{g}_z \rbrace$, the LiDAR frame denoted by $\mathcal{L}= \lbrace \bm{c}_\mathcal{L}, \mathbf{l}_x, \mathbf{l}_y, \mathbf{l}_z \rbrace$; and the STD descriptor frame $\mathcal{T} = \lbrace \bm{c}_\mathcal{T}, \mathbf{t}_x, \mathbf{t}_y, \mathbf{t}_z \rbrace$

\section{Point Cloud Feature Extraction} \hypertarget{pc_feature_extraction}{} 
\label{sec:pc_feature_extraction}

\subsection{Edges and Surfaces}

LiDAR-only odometry determines the pose of a robot by analyzing consecutive LiDAR scans. The feature-based matching method extracts feature points from the LiDAR point cloud, which are used to estimate the transformation between consecutive scans. The extraction of these feature points is based on the work presented in \cite{velasco2023lilo} and developed in the \texttt{pc\_feature}\footnote{\href{https://github.com/AUROVA-LAB/aurova_preprocessed/tree/master/pc_features}{https://github.com/AUROVA-LAB/aurova\_preprocessed/\\\hspace*{2.2em}tree/master/pc\_features}} repository. In this process, edges, surfaces, and ground are identified from a point cloud. To achieve this, the point cloud is first converted into a spherical range image (SRI). This process allows us to convert $\{\mathbf{p}_j,~j=1,\cdots m\} \in \mathbb{R}^3$ LiDAR scans data into a planar spherical projection $\mathbf{s}_{j} \in \mathbb{R}^2$ representation. The subscript $j$ denotes a specific point within the point cloud, and $m$ represents the total number of points captured in each scan by the LiDAR sensor. Subsequently, convolutional filters based on the Sobel operator are applied to the SRI image to extract the scenario features.

\subsection{STD: Stable Triangle Descriptor}
\label{sec:std_descriptor}

STD descriptors, introduced in \cite{std}, serve as a set of global descriptors for 3D place recognition. This descriptor remains unchanged under translations and rotations, ensuring its invariance from any perspective. For this purpose, we start from the idea that a triangle in Euclidean space has a unique shape. The length of its sides and angles are invariant to any kind of rigid transformation. Based on this property, the first step is to accurately extract a collection of local key points that depict the vertices of these descriptors.
To extract these types of descriptors, the point cloud data set needs to be voxelized first. Next, based on the covariance from this dataset, one determines which points lie within a plane. This is achieved by starting from a set of nearby points \mbox{$\mathbf{A} \in \{\mathbf{p}_j,~j=1,\cdots m\}$}, and their normal vector $\mathbf{n}$ is calculated. The set of points with a similar normal within a given threshold are considered as elements of the same vertex. Thus, each vertex $\{\mathbf{v}_1,\mathbf{v}_2,\mathbf{v}_3\}$ of the STD descriptor is defined as the midpoint $\bm{c}$ of this set of points sharing a similar normal within the threshold.

Next, the distance between each pair of these vertices is computed to establish an ordered sequence of the local points, ensuring a consistent reference system for these descriptors. Given the sides of the triangle in the STD descriptor, $l_{12}, l_{23}, l_{13}$, the rule that must be followed is: $l_{12} \leq l_{23} \leq l_{13}$.

\begin{figure}[ht]
    \centering
        \subfloat[][STD elements \label{fig:std_componets}]{
        \includegraphics[width=0.55\columnwidth]{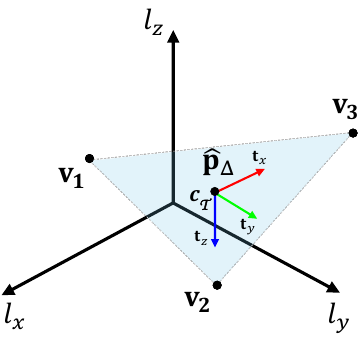}}    
    \\
        \subfloat[][Frame of each STD \label{fig:std_reference_system}]{%
        \includegraphics[width=0.9\columnwidth]{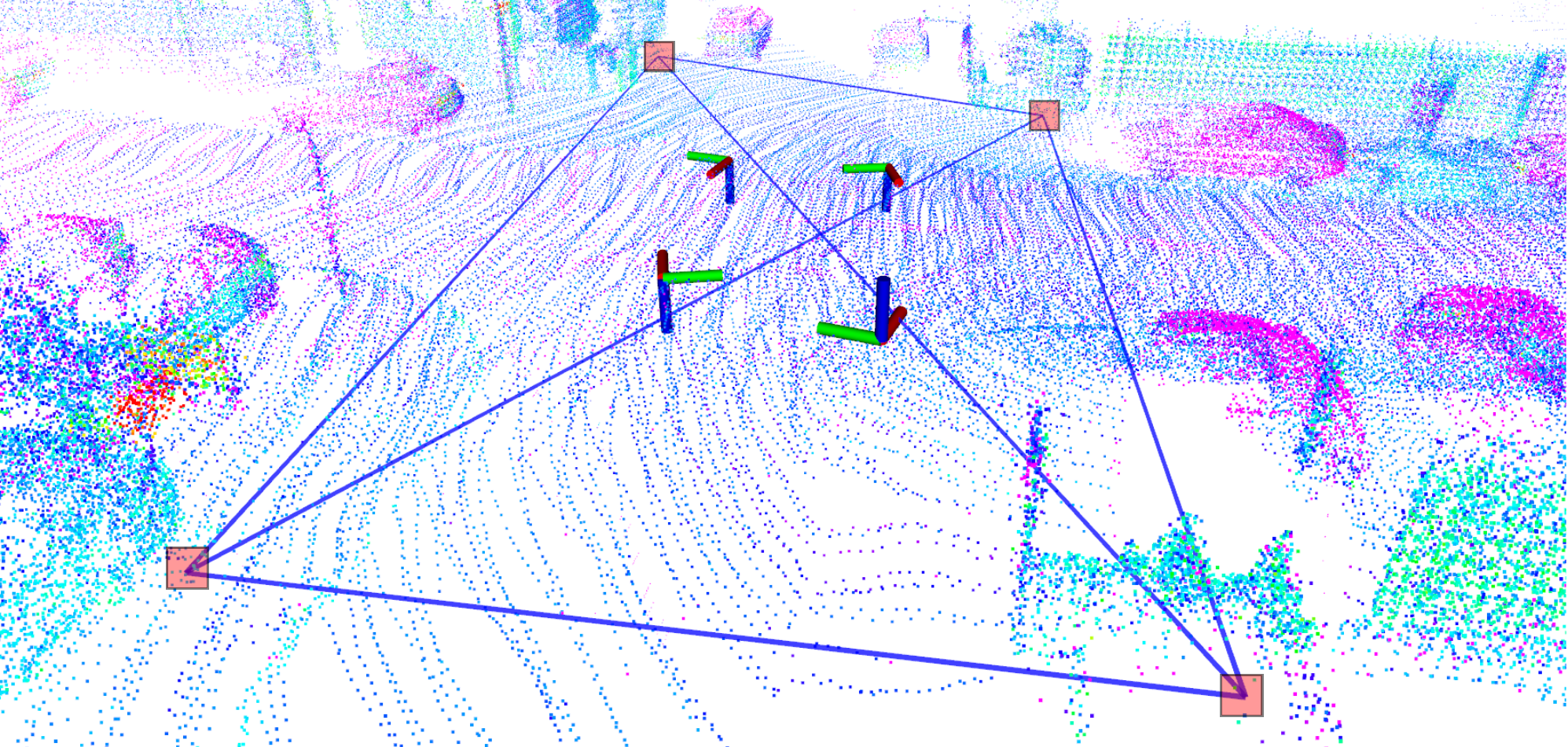}}   
        \caption{(a) The components of STD descriptor. (b) Reference frame of four STD descriptors detected in a point cloud. Each triangle has a unique reference system that does not vary with rigid transformations.}
\end{figure}

In addition to these elements, in this article, we have introduced a new feature to the approach of STD descriptors, which consists of generating a reference system denoted by $\mathcal{T}= \lbrace \bm{c}_\mathcal{T}, \mathbf{t}_x, \mathbf{t}_y, \mathbf{t}_z \rbrace$ located in the centroid $\bm{c}_\mathcal{T}$ of the vertices of each STD frame, and the orientation angles are determined as follows: \mbox{$\mathbf{t}_{x} = \frac{\mathbf{v}_3 - \bm{c}_\mathcal{T}}{\| \mathbf{v}_3 - \bm{c}_\mathcal{T}\|}$};~\mbox{$\mathbf{t}_{y} = \frac{\mathbf{v}_2 - \mathbf{v}_1}{\| \mathbf{v}_2 - \mathbf{v}_1 \|}$};\mbox{$~\mathbf{t}_{z} = \mathbf{t}_{x} \times \mathbf{t}_{y}$}. Then, the orthogonality of the $\mathbf{t}_{y}$  must be corrected using \mbox{$\mathbf{t}_{y} = \mathbf{t}_{z} \times \mathbf{t}_{x}$}.

Thus, we define the STD descriptor extracted from the LiDAR scans as $\{^\mathcal{L} \hat{\mathbf{p}}_{\Delta_j}, ~j=1,\cdots n \} \in \mathcal{H}$;  where $n$ is the number of STD descriptors in a $k$-th scan of the LiDAR. In this way, a reference system can be generated for each of the STD descriptors $^\mathcal{L}\hat{\mathbf{p}}_{\Delta}$, as can be seen in Fig. \ref{fig:std_reference_system}.

In \cite{LTAOM2024}, the STD descriptor is used to correct loop closure errors in high-drift odometry systems. This is achieved by generating a map of STD descriptors, which are organized in a hash table \cite{hashtable} for quick search. In our research, the aim is not to correct such drift errors but to improve the accuracy of odometry in each estimation. For this, we have generated a new method for generating STD maps (see section \ref{subsec:localmaps}). This is achieved by accumulating the STD descriptors detected in the scenario with their respective estimated pose with respect to a global reference system.

\section{Pose Estimation}
\hypertarget{Pose estimation}{} 
\label{section:Pose estimation}
 LOAM methods based on matching-of-features, align current features with local feature maps respectively for pose estimation. Usually, the features extracted from the point cloud are edges and surfaces \cite{wang2021floam, velasco2023lilo, legoloam2018, xu2021fast}. In addition, the alignment methods between the current features and their local map are performed using nearest neighbor search methods \cite{cai2021ikd, blanco2014nanoflann}. In our approach, in addition to using edge $^\mathcal{L}\mathbf{P}_e$ and surface $^\mathcal{L}\mathbf{P}_s$ features, we employ STD descriptors $^\mathcal{L}\hat{\mathbf{P}}_{\Delta}$ extracted from the point cloud $^\mathcal{L}\mathbf{P}$. The inclusion of STD descriptors allows us to use the reference frame of each descriptor containing both position and attitude for pose estimation. Thus, in addition to the edge and surface feature maps $^\mathcal{G}\mathbf{P}_e$, $^\mathcal{G}\mathbf{P}_s$ for pose estimation, we use a current measurement of descriptors $^\mathcal{G}\hat{\mathbf{P}}_{\Delta}$ at $k$-th scan of the LiDAR.  More details on the construction of these features maps can be found in section \ref{subsec:localmaps}. Once these features and descriptors are extracted, an optimization process is employed to minimize the transformation between the current features and the map. This optimization aims to find the transformation that best aligns the current features $^\mathcal{L}\mathbf{P}_{e}, ^\mathcal{L}\mathbf{P}_{s}$, and the $^\mathcal{L}\hat{\mathbf{P}}_{\Delta}$ descriptors with the corresponding features and descriptors in the local map in frame $\mathcal{G}$.

\subsection{Dual-Quaternions Parametrization}
\hypertarget{DQ_Parametrization}{} 
There are several representations of $SE(3)$, including Euler angles, quaternions, and homogeneous matrix representations, among others \cite{gallo2022so}. Each of these can represent the rotation and translation of a rigid body. On the other hand, dual quaternions have the advantage of compactly representing both elements of a rigid body \cite{duales_quat} \cite{sola2018micro}. Additionally, several geometric primitives such as points, planes, lines, spheres, and coordinate systems can be easily described. Furthermore, dual quaternions are easy to map into a vector structure, facilitating the generation of optimization algorithms for pose estimation. 

In this way, the descriptors described in the section \ref{sec:pc_feature_extraction} can be represented by the isomorphic algebra of the dual quaternion.

\begin{figure}[ht]
    \centering
        \subfloat[][Plücker Line \label{fig:plucker_line}]{
        \includegraphics[width=0.48\columnwidth]{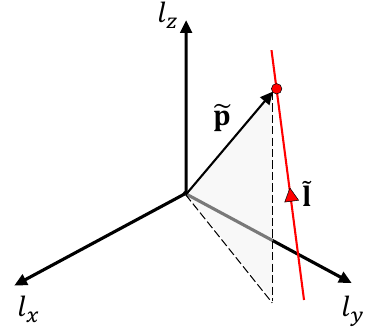}}    
    \hfill    
        \subfloat[][Plane  \label{fig:plano_dual}]{%
        \includegraphics[width=0.48\columnwidth]{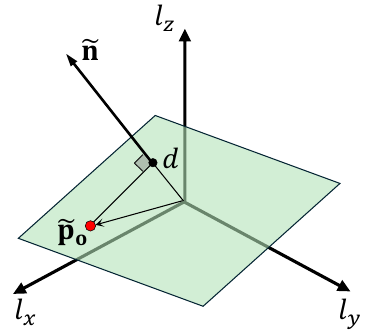}}    
    \\
        \subfloat[][STD Descriptor \label{fig:std_dual}]{%
        \includegraphics[width=0.5\columnwidth]{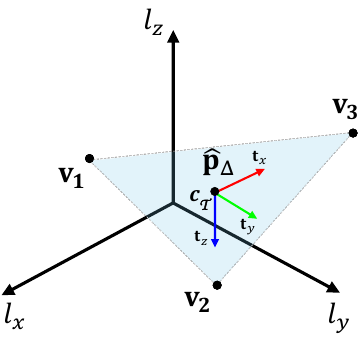}}   
    \caption{Elements extracted from the input point cloud parameterized as dual quaternions.}
\end{figure}

\subsubsection{Plücker Lines}
\label{subsec:plucker}
Plücker coordinates are used to represent lines in three-dimensional space \cite{adorno2020dq,jia2020plucker}. A Plücker line has four degrees of freedom: three for the translation and one for the rotation of the line around its own axis. To represent this type of line, a direction vector $\tilde{\mathbf{l}} \in \mathbb{H}_p$ passing through a point $\tilde{\mathbf{p}} \in \mathbb{H}_p$ is necessary. Plücker coordinates are defined by $(\tilde{\mathbf{l}}, {\tilde{\mathbf{m}}})$, where $\tilde{\mathbf{m}} = \tilde{\mathbf{p}} \times \tilde{\mathbf{l}}$.

Thus, a line in three-dimensional space can be represented by a dual quaternion and its definition as Plücker coordinates (see Fig. \ref{fig:plucker_line}). A Plücker line is defined as \eqref{eq:plucker_line}.

\begin{equation}
    ~~~~~~~~~~~~~~~~~~~~\hat{\mathbf{l}} = \tilde{\mathbf{l}} + \epsilon\tilde{\mathbf{m}}
    \label{eq:plucker_line}
\end{equation}

\subsubsection{Planes represented by dual quaternion}
\label{subsec:planes_dq}

A plane can be described with the algebra of dual quaternion by its normal vector $\tilde{\mathbf{n}} \in \mathbb{H}_p$ and a perpendicular distance $d$ given by \mbox{$d = \tilde{\mathbf{n}} \cdot \tilde{\mathbf{p}}_\mathbf{o}$}, where $\tilde{\mathbf{p}}_\mathbf{o} \in \mathbb{H}_p$ is any point on the plane represented as a pure quaternion . The Fig. \ref{fig:plano_dual} shows the elements that constitute a plane for its representation as a dual quaternion.

Thus, a plane can be represented in dual space as follows:

\begin{equation}
    ~~~~~~~~~~~~~~~~~~~\hat{{\bm{\pi}}}  = \tilde{\mathbf{n}} + \epsilon( \tilde{\mathbf{p}}_\mathbf{o}\cdot \tilde{\mathbf{n}})
    \label{eq:plano_dual}
\end{equation}

\subsubsection{STD descriptors represented by unit dual quaternions}

The reference system of STD descriptors consists of an orientation and a translation vector. Both elements can be easily represented as unit dual quaternion. In Figure \ref{fig:std_dual}, the centroid $\bm{c}_\mathcal{T}$ of the descriptor is shown, where the descriptor's reference system is represented.

Finally, we can represent the descriptor with a unit dual quaternion $\hat{\mathbf{p}}_{\Delta}$ as follows:

\begin{equation}
    \hat{\mathbf{p}}_{\Delta} = \mathbf{\tilde{q}}_{\mathbf{r}} + \epsilon\frac{1}{2} \left ( \mathbf{\tilde{q}}_{\mathbf{t}}\otimes\mathbf{\tilde{q}}_{\mathbf{r}} \right )
    \label{eq:std_dual_represetnacion}
\end{equation}

Where $\mathbf{\tilde{q}}_{\mathbf{r}} \in \mathbb{S}^3$ is the orientation and $\mathbf{\tilde{q}}_{\mathbf{t}} \in \mathbb{H}_p $ is the translation of the descriptor, respectively.

\subsection{Residuals Computation}
\hypertarget{residuals}{} 
By representing surfaces, edges, and STD descriptors as dual quaternions, we can create cost functions aimed at minimizing errors through transforming each feature to its corresponding local map using the unit dual quaternion ${\mathbf{\hat{q}}}^{op}_{k}$. The Fig. \ref{fig:pipeline_DualQuat-LOAM}  illustrates the elements used in the point-to-edge, point-to-plane and STD cost function within the residuals computation.

\subsubsection{Point-to-edge Cost Function}

A cost function \eqref{eq:costo_bordes_dual} is defined that minimizes the edge error $\alpha^{e}_k$ when transforming an edge point $\mathbf{p}_e \in \mathbb{R}^3$, belonging to the set of edges in a point cloud $^\mathcal{L}\mathbf{P}_{in}$ of a LiDAR sensor scan, towards an edge in the local map $^\mathcal{G}\mathbf{P}_e$. 

\begin{equation}
    \begin{aligned}    
    \mathbf{\hat{q}}^\varepsilon_{e} &= \mathbf{\hat{q}}_{k} \boxtimes \hat{\mathbf{p}}_e \boxtimes  \mathbf{\hat{q}}_{k}^{3*} \\
    \alpha^{e}_k &=  \sum \left \|   \left (  \mathcal{D}\left (\mathbf{\hat{q}}^\varepsilon_{e}\right )\times \hat{\mathbf{l}}\right ) - \tilde{\mathbf{m}} \right \|         
    \end{aligned}
\label{eq:costo_bordes_dual}
\end{equation}

Where $\hat{\mathbf{p}}_e \in 
\mathcal{H}$ is the representation of the point $\mathbf{p}_e \in \mathbb{R}^3$ as a dual quaternion using the  Plücker line  formulation
(see section \ref{subsec:plucker})
.
\subsubsection{Point-to-Plane Cost Function} 

Similar to the edge, a cost function \eqref{eq:costo_plano_dual} is determined that minimizes the surface error $\alpha^{s}_k$ when transforming a surface point $\mathbf{p}_s$ belonging to the surface group of a point cloud $^\mathcal{G}\mathbf{P}_{in}$ of a LiDAR sensor scan transformed to the global reference frame, towards a plane in the local map $^\mathcal{G}\mathbf{P}_\mathcal{s}$.

\begin{equation}
    \begin{aligned}    
    \mathbf{\hat{q}}^\varepsilon_{s} &= \mathbf{\hat{q}}_{k} \boxtimes \hat{\mathbf{p}}_s \boxtimes  \mathbf{\hat{q}}_{k}^{3*} \\
    \alpha^{s}_k &=  \sum \left (  \left ( \mathcal{D}\left (\mathbf{\hat{q}}^\varepsilon_{s}\right )\otimes \tilde{\mathbf{n}}\right) - d  \right )    
    \end{aligned}
\label{eq:costo_plano_dual}
\end{equation}

Where $\hat{\mathbf{p}}_s \in \mathcal{H}$ represents the point $\mathbf{p}_{s} \in \mathbb{R}^3$ as a dual quaternion through a plane in dual quaternions (refer to section \ref{subsec:planes_dq}).

\subsubsection{STD Descriptor Cost Function}

Since the STD descriptors contain both translation and orientation of a reference system, we can represent the descriptor with a unit dual quaternion. In this way, we define the cost function that estimate the dual quaternion $\hat{\mathbf{q}}_{k}$ that transforms the current STD descriptor $\hat{\mathbf{p}}_{\Delta} $ to its corresponding one $\mathbf{_m\hat{p}}_{\Delta} $ in the STD map $^\mathcal{G}\hat{\mathbf{P}}_ {\Delta}$. 

\begin{equation}
    \begin{aligned}    
    \mathbf{\hat{q}}^{\varepsilon}_{\Delta} &= \mathbf{_m\hat{p}}_{\Delta}^{2*} \boxtimes \hat{\mathbf{q}}_{k} \boxtimes  \mathbf{\hat{p}}_{\Delta} \\
    \alpha^{\Delta}_k& = \sum||\mathbf{\hat{q}}_{u}-\mathbf{\hat{q}}^{\varepsilon}_{\Delta} ||    
    \end{aligned}
\label{eq:costo_std_dual}
\end{equation}

Where  $\mathbf{\hat{q}}_{u}$ is defined as the identity element $\mathbf{\hat{q}}_{u} = \mathbf{1} + \epsilon \mathbf{0}$, which represents no change in the geometric structure of dual quaternion $\mathcal{H}$.

Thus, the optimal dual transformation quaternion $\bm{\mathbf{\hat{q}}^{op}_{k}}$ that minimizes each feature and descriptor error is defined in \eqref{eq:cost_dual}.

\begin{equation}
     \bm{\mathbf{\hat{q}}^{op}_{k}}= \operatorname*{argmin}_{\mathbf{\hat{q}}_{k}}  \left ( \alpha^{e}_k + \alpha^{s}_k +\alpha^{\Delta}_k \right)
       \label{eq:cost_dual}   
\end{equation}
\subsection{Dual-Quaternions Optimization}
\hypertarget{DQ_optimization}{} 
Dual quaternions provide a compact and efficient representation of rigid body transformations, combining both rotation and translation into a single mathematical framework. Optimizing in the space of dual quaternions $\mathcal{H}$ requires specialized techniques that can handle the non-Euclidean nature of this space. 

\begin{figure}[ht]
    \centering
    \includegraphics[width=0.8\columnwidth]{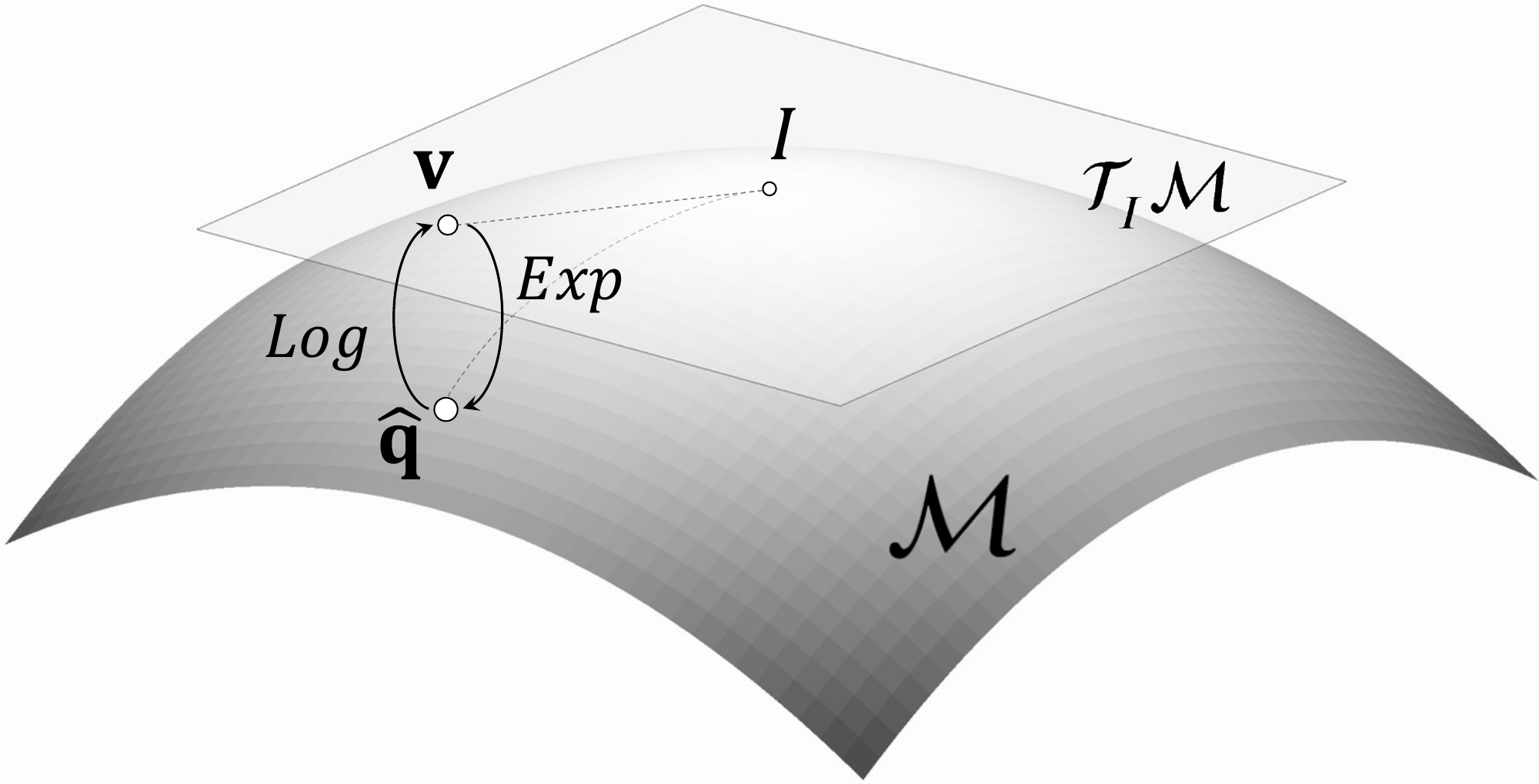}
    \caption{Illustrative representation of the dual quaternion \textit{Manifold} $\mathcal{M}$ and its tangent plane $\mathcal{T}_{I}\mathcal{M}$}
    \label{fig:manifold_dual}
\end{figure}

Ceres is an open-source library designed for solving non-linear least squares problems using Gauss-Newton techniques \cite{Agarwal_Ceres_Solver_2022}. This library is widely used in applications such as camera calibration, 3D reconstruction, and pose estimation in navigation systems. In this research, DualQuat-LOAM have been developed using the Ceres optimizer. Ceres' flexibility in creating non-Euclidean \textit{Manifolds} makes it ideal for applications that require optimization involving quaternions or dual quaternions. In DualQuat-LOAM, we have developed a custom \textit{Manifold} in Ceres to optimize dual quaternions directly and is illustrated in Fig. \ref{fig:manifold_dual}. This allows us to efficiently formulate and solve optimization problems using dual quaternions. By leveraging the capabilities of Ceres, we can perform optimizations that respect the mathematical properties of dual quaternions. Due to the fact that optimizers work by numerical methods operating in $\mathbb{R}^n$, it is necessary to perform a mapping between the geometric structure $\mathcal{H}$ where the values of dual quaternions are estimated.The Exp and Log functions enable the mapping $\mathbb{R}^n \leftrightarrow \mathcal{H}$, which maintains the geometric structure of dual quaternions and facilitates the application of optimization algorithms within Euclidean space.

\subsubsection{\textit{Plus} and \textit{Minus} Functions in Ceres}

\underline{The \textit{Plus} function} defines how to update parameters within the \textit{Manifold} $\mathcal{M}$. It is used to apply an increment (delta $\delta$) to the current parameters $\boxplus (\mathbf{\hat{q}},\delta)$. This function computes the displacement along $\delta$ in the tangent space at $\mathbf{v}$, and then projects it back onto the \textit{Manifold}  in $\mathbf{\hat{q}}$.    

\underline{The \textit{Minus} function}, denoted as $\boxminus (\mathbf{\hat{q}}_2,\mathbf{\hat{q}}_1)$, computes the difference between two points on the \textit{Manifold} and projects this difference onto the tangent space.

Thus, for a \textit{Manifold} parameterized in dual quaternions, we define the $\boxplus$ and $\boxminus$ functions responsible for mapping to the increment vector as follows:

\begin{equation}
\begin{aligned}
\mathbf{\hat{q}}_1\boxplus \mathbf{v}_1& = \text{Exp}(\mathbf{v}_1)\boxtimes\mathbf{\hat{q}_1}\\
    \mathbf{\hat{q}}_2\boxminus \mathbf{\hat{q}}_1& = \text{Log}(\mathbf{\hat{q}}_1^{2*}\boxtimes\mathbf{\hat{q}}_2)
\end{aligned}    
    \label{eq:dual_plus_minus}
\end{equation}

\section{Map Update}
\hypertarget{mapping}{} 
\label{section:mapping}

Once the state $ \bm{\mathbf{\hat{q}}^{op}_{k}}$ is calculated, which contains the estimated pose, the local feature maps are generated for each feature point and descriptor. These points are projected onto the LiDAR frame $\mathcal{L}$, and then transformed into the map on the global frame $\mathcal{G}$ by accumulating the features $^\mathcal{L}\mathbf{P}_{e_k}$, $^\mathcal{L}\mathbf{P}_{s_k}$, and $^\mathcal{L}\mathbf{\hat{P}}_{\Delta_k}$ of the LiDAR sensor frame at $k$-th scan. These features points and descriptors are added to the existing  map, which contains the points obtained from previous feature and descriptor extraction steps. 

On the other hand, the global point cloud map is updated after keyframes when a threshold of position or translation movement is exceeded. Finally, this global point cloud map is downsampled using a 3D voxel grid approach to prevent memory overflow.

\subsection{Local Map}

The local map serves as a cumulative reference where current features and descriptions are aligned and combined with those already present, thus enabling a pose estimation generating system odometry. The process ensures that the newly captured features and descriptors are related to the already established spatial structure, thereby producing the motion estimate.

\subsubsection{Edge and Surfaces Map}
As with LOAM-based methods, each feature point projected to the LiDAR reference frame is transformed to the global frame using the previously estimated poses \cite{zhang2017loam,wang2021floam,velasco2023lilo, xu2021fast}.  The difference is that, in our proposal, the projection is computing by means of the algebra of dual quaternions. Due to the fact that, both edges and surfaces are features that belong to a pure dual quaternion $\mathbb{H}_p$, the transformation between the LiDAR $\mathcal{L}$ frame to the global frame $G$ is expressed as: $^\mathcal{G}\mathbf{\hat{p}}_{e_i} = \mathbf{^\mathcal{G}\hat{q}}_{k-1} ~\boxtimes ~^\mathcal{L_k}\hat{\mathbf{p}}_{e_i} ~\boxtimes  ~\left(\mathbf{^\mathcal{G}\hat{q}}_{k-1}^{op}\right)^{3*}$ for edges,  and  $^\mathcal{G}\mathbf{\hat{p}}_{s_i} = \mathbf{^\mathcal{G}\hat{q}}_{k-1} ~\boxtimes ~^\mathcal{L_k}\hat{\mathbf{p}}_{s_i} ~\boxtimes  ~\left(\mathbf{^\mathcal{G}\hat{q}}_{k-1}^{op}\right)^{3*}$ for surface; where $i$ is the number of features in $k$-th LiDAR scan.  

\subsubsection{STD Map}
\begin{figure}[ht]
    \centering
    \includegraphics[width=1.0\linewidth]{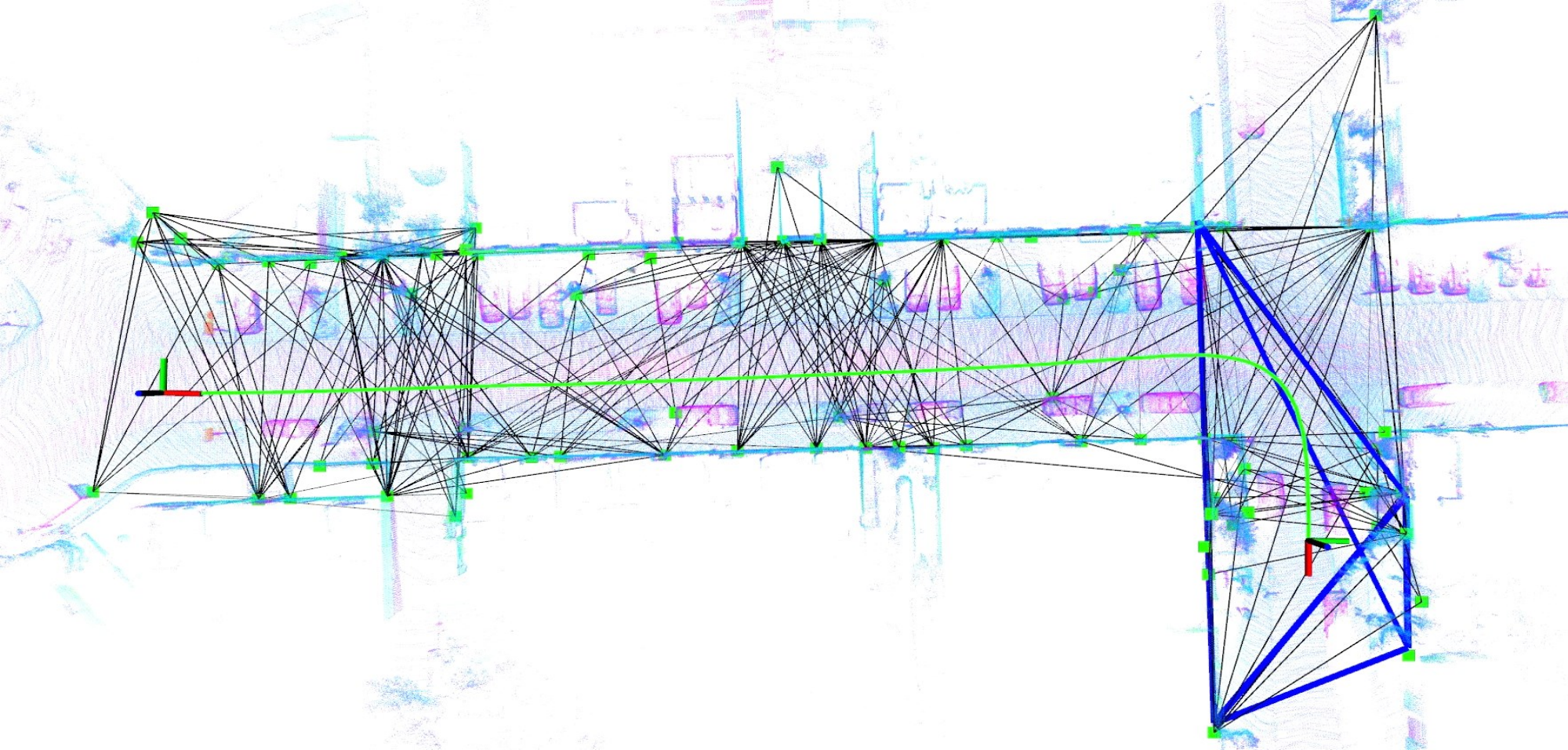}    
    \caption{Current STD descriptors and the map of STD descriptors represented as blue triangles and black triangles, respectively.}
    \label{fig:std_curr_map}
\end{figure}

The STD map descriptors $^\mathcal{G}\hat{\mathbf{P}}_{\Delta}$ is defined as the accumulation of descriptors in a data window of size $h$. In this way, each STD descriptor at $k$-th LiDAR scan is transformed by:  $^{\mathcal{G}}\hat{\mathbf{p}}_ {\Delta_i}~={^\mathcal{G}\hat{\mathbf{q}}^{op}_{k-1}} ~\boxtimes~ ^{\mathcal{L}_k}\hat{\mathbf{p}}_ {\Delta_i}$; where  $^{\mathcal{G}}\hat{\mathbf{p}}_ {\Delta_i}$ contains the $i$ descriptors $^{\mathcal{L}_k}\hat{\mathbf{p}}_ {\Delta_i} ; [~i-1, i-2, \ldots, i-h~]$. In order to reduce the number of STD descriptors, we implemented a filter to optimize the accumulated descriptor density. This process involves clustering nearby vertices within a specific radius around each vertex of the triangle in the STD map $^\mathcal{G}\hat{\mathbf{P}}_ {\Delta}$. By unifying adjacent vertices, redundancy in the representation of STD descriptors is avoided, thus reducing the over-accumulation of descriptors and improving the construction of the STD map. Fig. \ref{fig:std_curr_map} shows an example of the current STD descriptors $^{\mathcal{L}_k}\hat{\mathbf{p}}_ {\Delta_i}$ represented as blue triangles in the LiDAR reference frame, and the STD descriptor map  $^\mathcal{G}\hat{\mathbf{P}}_ {\Delta}$ is shown as black triangles. In this figure, only four current STD descriptors and an STD map data window of $h=50$ elements are used to ensure clear and understandable visualization. For this representation, the first $14.5~[s]$ of the 00 sequence of the KITTI dataset has been used. 

For the matching between the current STD descriptors  transformed to the global frame and the accumulated  STD map  $^\mathcal{G}\hat{\mathbf{P}}_ {\Delta_{M}}$, we employed a K-Nearest Neighbors (KNN) algorithm \cite{blanco2014nanoflann}, which matches the elements of each descriptor (see section \ref{sec:std_descriptor}). This approach allows us to identify and relate the descriptors corresponding to the generated map. 

\begin{figure}[ht]
    \centering
    \includegraphics[width=1.0\linewidth]{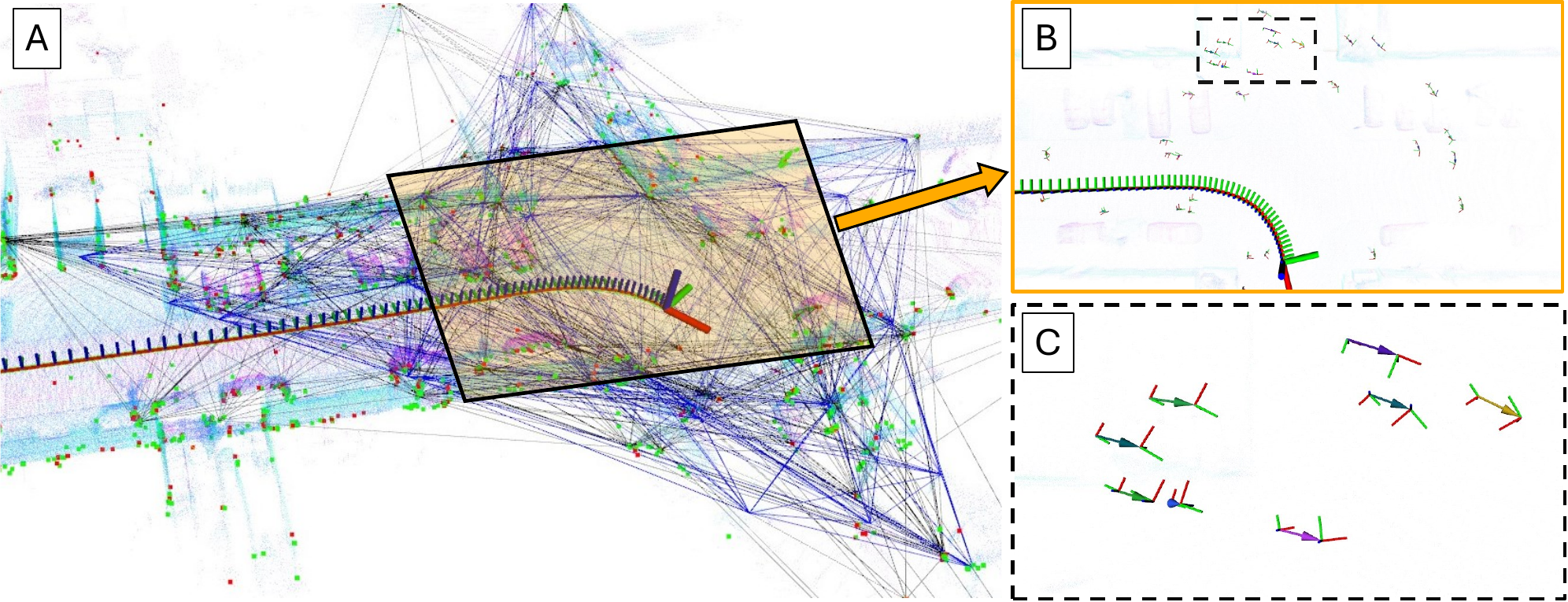}    
    \caption{Matching process between the current STD descriptors and the STD map. (A) Sequence 00 of the KITTI dataset with the STD map. (B) and (C) show the matching of the STD descriptors and the STD map, the direction of the arrows show the same direction of the LiDAR sensor movement.}    
    \label{fig:std_mapping}
\end{figure}

The Fig. \ref{fig:std_mapping} shows the matching process using the K-NN algorithm. Fig. \ref{fig:std_mapping}.A presents the scenario in which the STD descriptors have been extracted and the STD map is being generated. In Fig. \ref{fig:std_mapping}.B, the direction vectors indicating how the current STD descriptors correlate to the STD map are shown; in addition, the axes of each triangle corresponding to both the current and map STD descriptors are highlighted in the Fig. \ref{fig:std_mapping}.C. 

\begin{figure}[ht]
    \centering
    \includegraphics[width=1.0\linewidth]{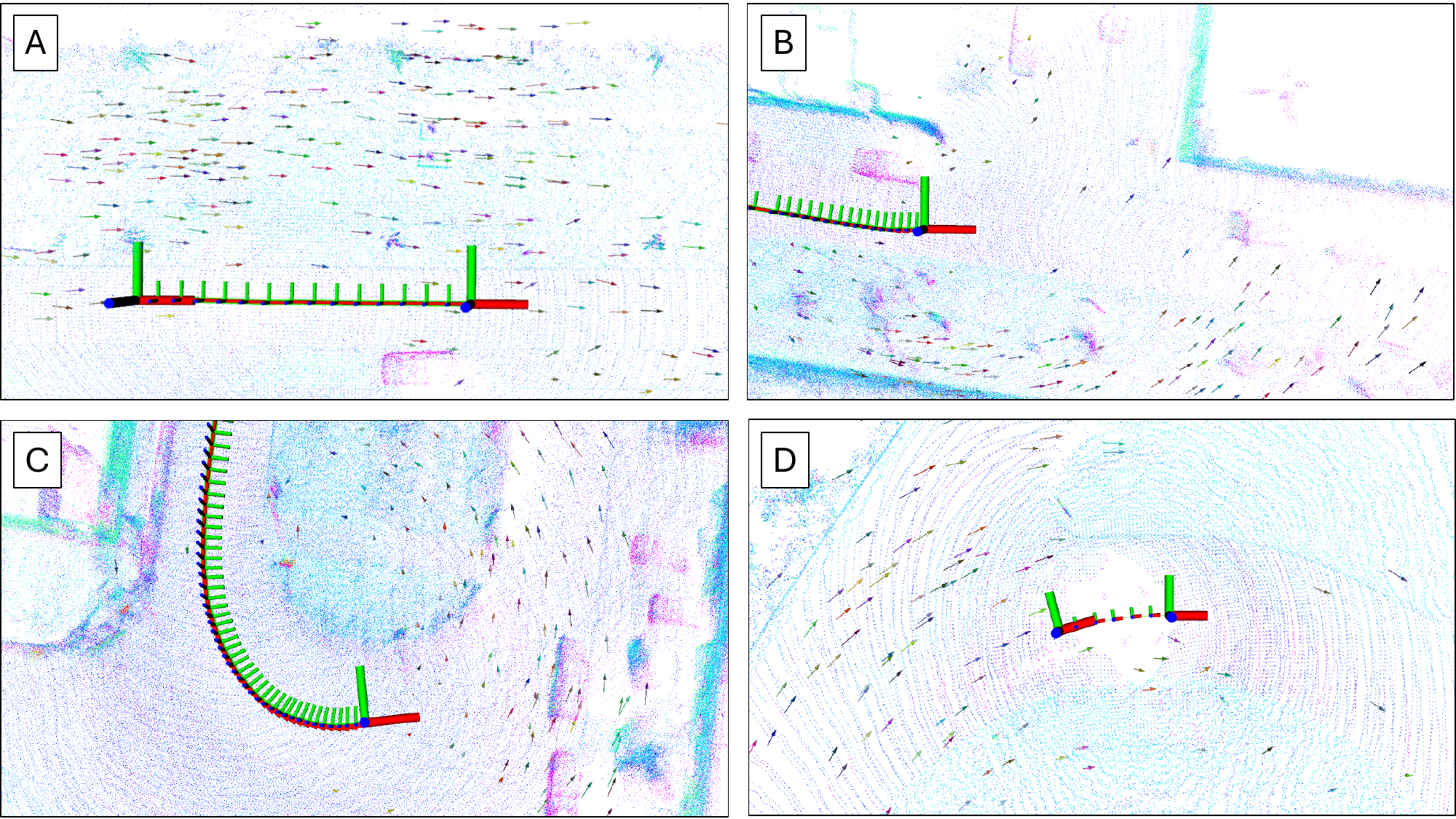}       
    \caption{Correspondence arrows from the STD map to the current STD descriptor. Examples executed in KITTI sequences.}  
    \label{fig:std_matching_map}
\end{figure}

Fig. \ref{fig:std_matching_map} shows examples of the correspondence between the current STD descriptors and the STD map, visualized by arrows indicating the match between them. The images correspond to different sequences of the KITTI dataset, with their respective timestamps. The sub-figures Fig. \ref{fig:std_matching_map}.A and Fig. \ref{fig:std_matching_map}.B are part of the sequence 00 at $1.56~[s]$ and $42.6~[s]$ respectively. The Fig. \ref{fig:std_matching_map}.C is the sequence 06 at $31.9~[s]$, and the Fig. \ref{fig:std_matching_map}.D is the sequence 01 at $0.83~[s]$.

\label{subsec:localmaps}

\section{Experiments and Results}

The experiments were performed on a computer equipped with an Intel® Core™ i5-13600KF processor and 16 GB of RAM. To evaluate the performance of our approach compared to other state-of-the-art methods, we used the KITTI dataset, selecting several LiDAR odometry methods that, like our approach, exclusively use point clouds for pose estimation, without the integration of other sensors. Our method, DualQuat-LOAM, was implemented in C++ and executed in ROS Noetic. The source code is available to the scientific community in the repository link\footnote{\href{https://aurova-projects.github.io/dualquat_loam/}{https://aurova-projects.github.io/dualquat\_loam/}}.
In addition, we have used the feature extraction method proposed in our previous work, LiLO \cite{velasco2023lilo}, and analyzed whether the parameterization of features into dual quaternions, together with the incorporation of STD descriptors, improves pose estimation. 

The configuration of the method was performed considering the following aspects. For edge and surface estimation, the 5 nearest neighbors were used to ensure accuracy in feature matching. For STD descriptors, 24 features were used, including vertices, normals, triangle side lengths and centroid, with the objective of capturing a detailed representation of geometric features and facilitating the matching between the current and map STD descriptors.

In terms of feature map management, $^\mathcal{G}\mathbf{P}_{e}$ edge and $^\mathcal{G}\mathbf{P}_{s}$ surface maps were accumulated within a $100 [m]$ along the LiDAR displacement axis, in order to optimize processing time during feature matching. The STD descriptor map was managed with a data window of $h = 10$, and with a maximum of 10 vertices per iteration, resulting in the generation of up to 120 triangles per iteration and a STD map composed of 1200 descriptors. In addition, filtering was applied to the vertices of the STD map using a radius of $0.2~[m]$, with the objective of grouping nearby vertices and avoiding redundancy in data accumulation.

\subsection{KITTI dataset evaluation}
\label{sec:KITTI dataset evaluation}
\begin{figure}[ht]
    \centering
    \includegraphics[width=1.0\columnwidth]{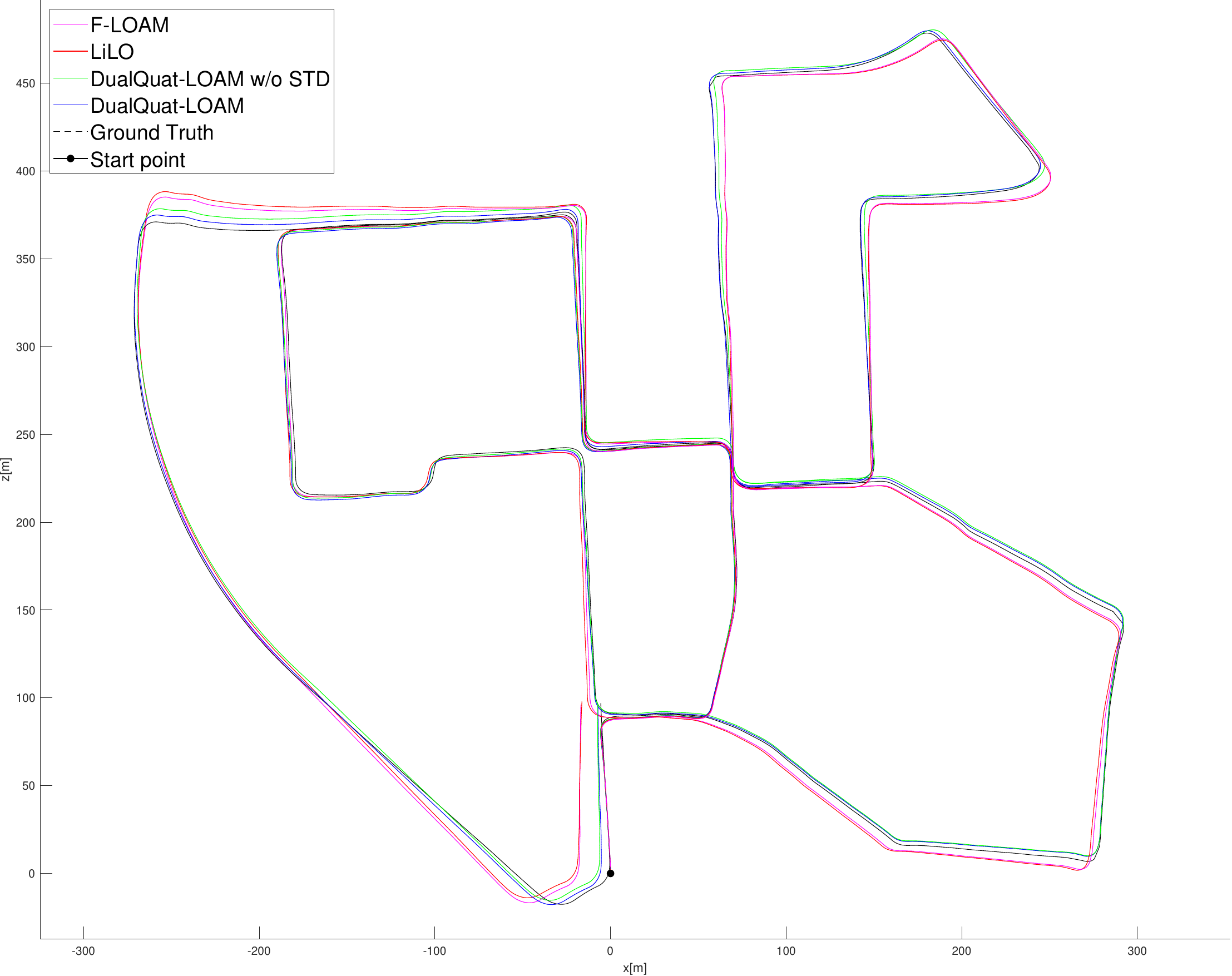}      
    \caption{Pose estimation comparison of different LiDAR odometry and mapping approaches with our system in the sequence 00 of the KITTI dataset.}    
    \label{fig:sequence00_comparative}
\end{figure}

To evaluate the performance of our DualQuat-LOAM approach, we use the KITTI dataset, which includes the ground truth of 11 different sequences covering urban environments, highways and country areas  \cite{KITTI_Benchmark}. For the evaluation of our method, we employ the tools described in \cite{zhan2019dfvo}, which allow us to calculate the average translation (ATE) and rotation (ARE) errors for each sequence.
First, we verify whether the contribution of STD descriptors and parameterization in dual quaternions improves pose estimation. For this purpose, we perform a comparison with the FLOAM and LiLO methods that use similar methodologies in terms of feature extraction and matching. Fig. \ref{fig:sequence00_comparative} presents the comparative results of the DualQuat-LOAM methods, both with and without STD descriptors, along with the LiLO and F-LOAM methods. The results show that the DualQuat-LOAM method with STD descriptors achieves a closer approximation to the ground truth in sequence 00 of the KITTI dataset. When comparing DualQuat-LOAM without STD descriptors with the LiLO method, it is observed that, although both employ the same features, the dual quaternion parameterization applied to the edge descriptors, surfaces and the optimizer reduces the drift error in the pose estimation by being closer to the ground truth.

\begin{table}[htbp]
   \centering
   \caption{KITTI dataset results. The values are represented as Translation(\%)/Rotation(º/100m). LOAM results were obtained from \cite{lodonet}, T-LOAM and F-LOAM from \cite{T-LOAM}, LEGO-LOAM and LiODOM from \cite{garcia2022liodom}, and LiLO from \cite{velasco2023lilo}. The best results are highlighted in red, and the second best in blue.}
    \resizebox{\columnwidth}{!}{%
    \begin{tabular}{lllllllll}
        \toprule
        \begin{tabular}[c]{@{}c@{}}\textbf{Seq.}\\\textbf{N°}\end{tabular} & \begin{tabular}[c]{@{}c@{}}\textbf{Path}\\\textbf{Len. (m)}\end{tabular} & \textbf{LOAM}& \textbf{T-LOAM } &  \textbf{F-LOAM} & \textbf{LiODOM} & \textbf{LiLO} & \begin{tabular}[c]{@{}c@{}}\textbf{DualQuat-}\\\textbf{LOAM}\end{tabular}  \\ 
        \hline
        \textbf{00}& 3714     & 0.78/0.53  & 0.98/0.60 & 1.11/\textcolor{blue}{0.40}  & 0.86/\textcolor{red}{0.35} & \textcolor{red}{0.71}/0.43 & \textcolor{blue}{0.75}/0.43\\
        \textbf{01}& 4268  & 1.43/0.55  & 2.09/0.52 & 3.01/0.85  & 1.30/\textcolor{red}{0.13} & \textcolor{blue}{1.29}/\textcolor{blue}{0.20} & \textcolor{red}{1.16}/0.21\\
        \textbf{02}& 5075.  & \textcolor{red}{0.92}/0.55  & 1.01/\textcolor{blue}{0.39} & 1.22/0.43  & 0.95/\textcolor{red}{0.31} & 1.06/0.39 & \textcolor{blue}{0.94}/0.42\\
        \textbf{03}& 563       & \textcolor{red}{0.86}/0.65  & \textcolor{blue}{1.10}/0.82 & 4.51/1.84  & 1.26/\textcolor{red}{0.23} & 1.22/\textcolor{blue}{0.31} & 1.30/0.36\\
        \textbf{04}& 397       & 0.71/0.50  & \textcolor{blue}{0.68}/0.68 & 0.93/0.63  & 1.41/\textcolor{red}{0.01} & 0.84/\textcolor{blue}{0.28} & \textcolor{red}{0.41}/0.40\\
        \textbf{05}& 2223      & 0.57/0.38  & \textcolor{blue}{0.55}/\textcolor{red}{0.32} & 0.63/\textcolor{blue}{0.32}  & 0.83/0.36 & \textcolor{red}{0.53}/0.34 & 0.56/0.38\\
        \textbf{06}& 1239     & 0.65/0.39  & 0.56/\textcolor{red}{0.31} & 2.15/0.74  & 0.83/0.33 & \textcolor{red}{0.54}/\textcolor{blue}{0.32} & \textcolor{blue}{0.54}/0.39\\
        \textbf{07}& 695      & 0.63/0.50  & \textcolor{red}{0.50}/\textcolor{blue}{0.47} & \textcolor{blue}{0.51}/\textcolor{red}{0.35}  & 0.88/0.61 & 0.60/0.61 & 0.62/0.63\\
        \textbf{08}& 3225  & 1.12/0.44  & \textcolor{blue}{0.94}/\textcolor{blue}{0.33} & 0.97/0.37  & \textcolor{red}{0.86}/\textcolor{red}{0.33} & 1.07/0.41 & 1.04/0.39\\
        \textbf{09}& 1717&  0.77/0.48  & 0.80/0.40 & 0.82/0.40  & 1.03/\textcolor{blue}{0.32} & \textcolor{red}{0.63}/\textcolor{red}{0.32} & \textcolor{blue}{0.68}/0.38\\
        \textbf{10}& 919& \textcolor{blue}{0.79}/0.57  & 1.12/0.61 & 2.52/0.96  & 1.20/0.29 & 0.99/\textcolor{blue}{0.33} & \textcolor{red}{0.72}/\textcolor{red}{0.33}\\
        \textbf{avg}&-        & \textcolor{blue}{0.84}/0.50  & 0.93/0.49 & 1.67/0.66  & 1.04/\textcolor{red}{0.30} & 0.86/\textcolor{blue}{0.36} & \textcolor{red}{0.79}/0.39\\
        \hline
    \end{tabular}%
    }
    \label{tab:dualquat_loam_comparativa}
\end{table}%

\begin{figure}
\centering
\newcommand\imgX{0.15}
\captionsetup[subfigure]{justification=centering}
     \begin{subfigure}[b]{\imgX\textwidth}
         \caption{01}
         \includegraphics[width=\textwidth]{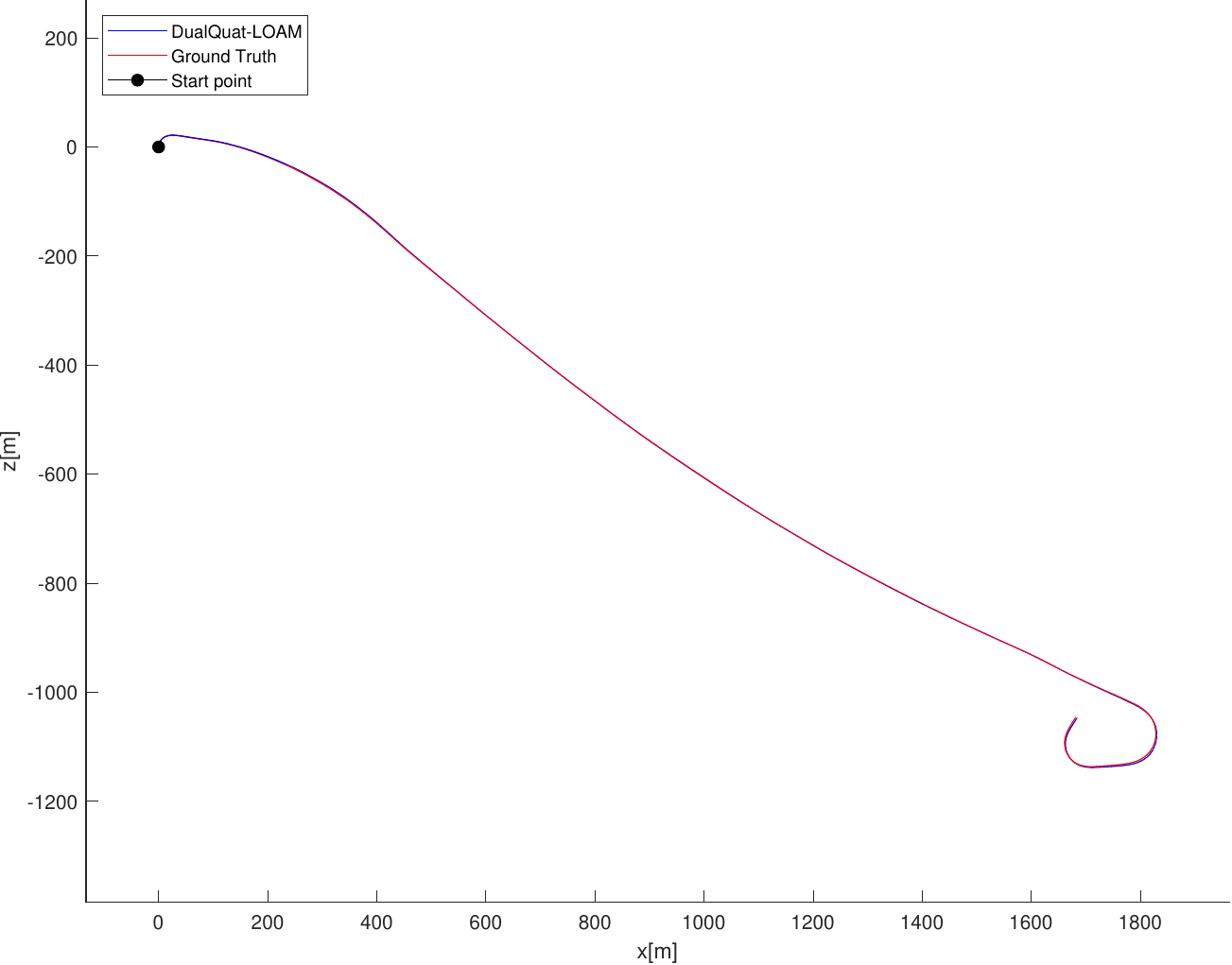}
         \label{fig:fig_kitti_01}
     \end{subfigure}
      \begin{subfigure}[b]{\imgX\textwidth}
         \caption{02}
         \includegraphics[width=\textwidth]{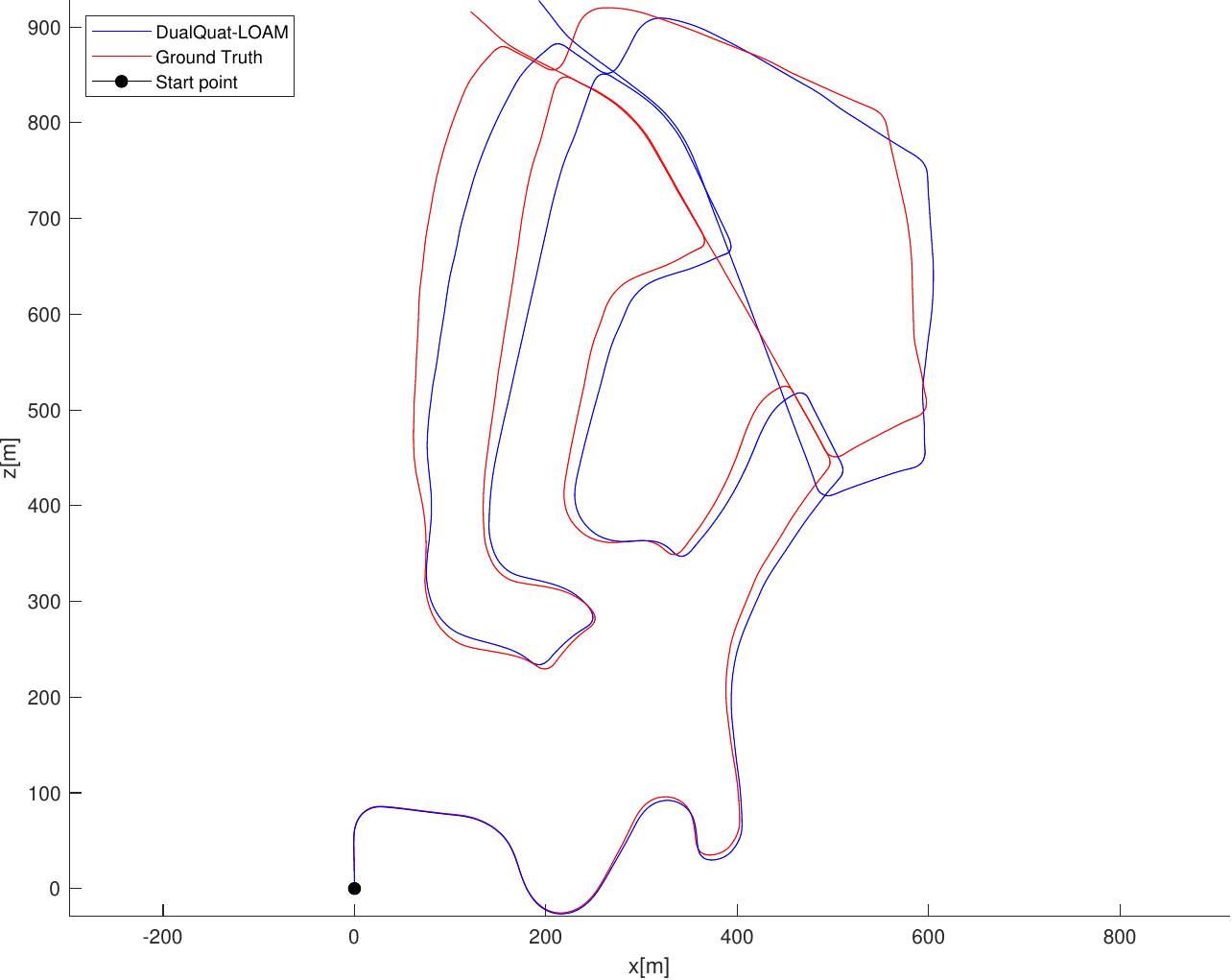}
         \label{fig:fig_kitti_02}
     \end{subfigure}
     \begin{subfigure}[b]{\imgX\textwidth}
         \caption{03}
         \includegraphics[width=\textwidth]{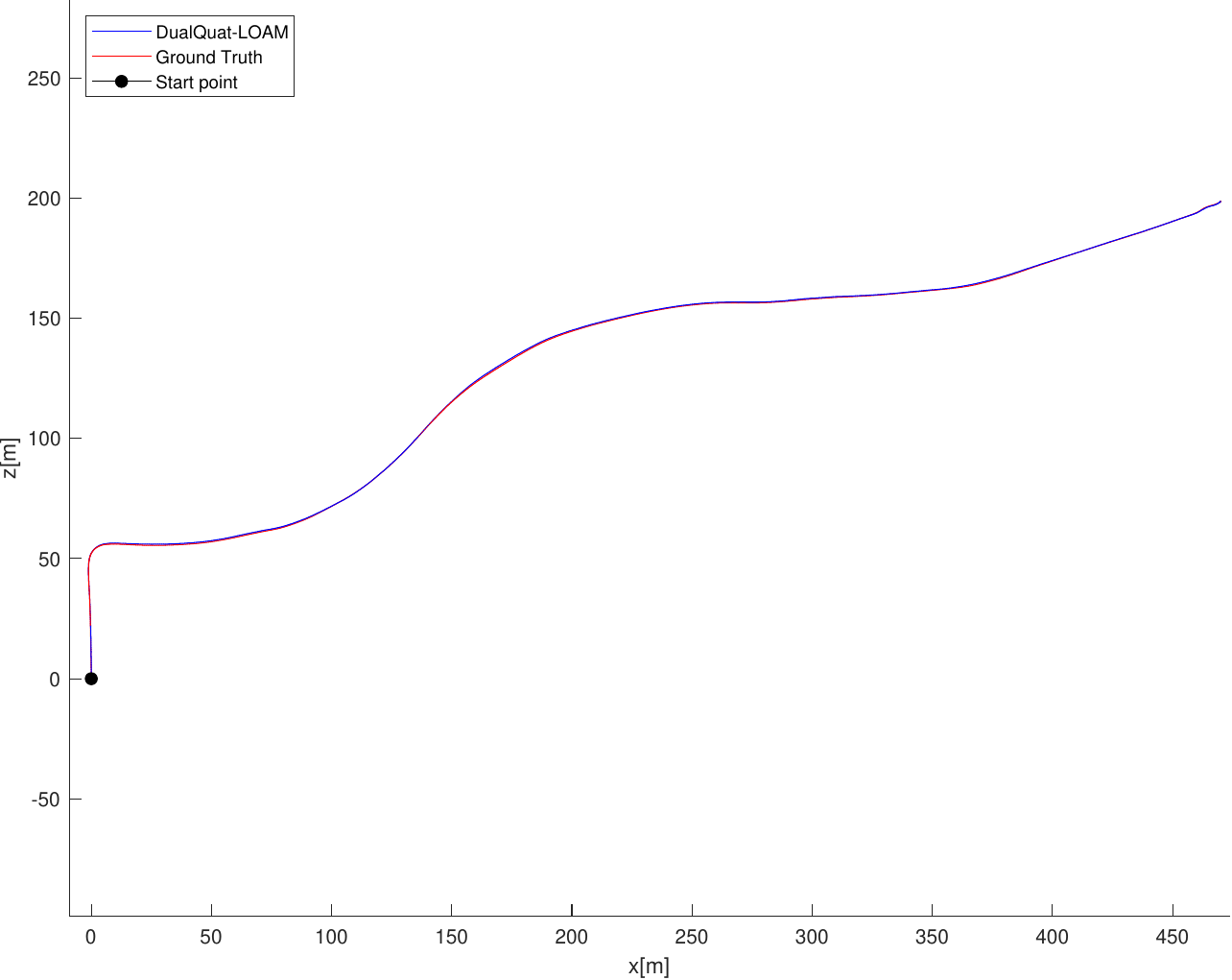}
         \label{fig:fig_kitti_03}
     \end{subfigure}
     \begin{subfigure}[b]{\imgX\textwidth}
         \caption{04}
         \includegraphics[width=\textwidth]{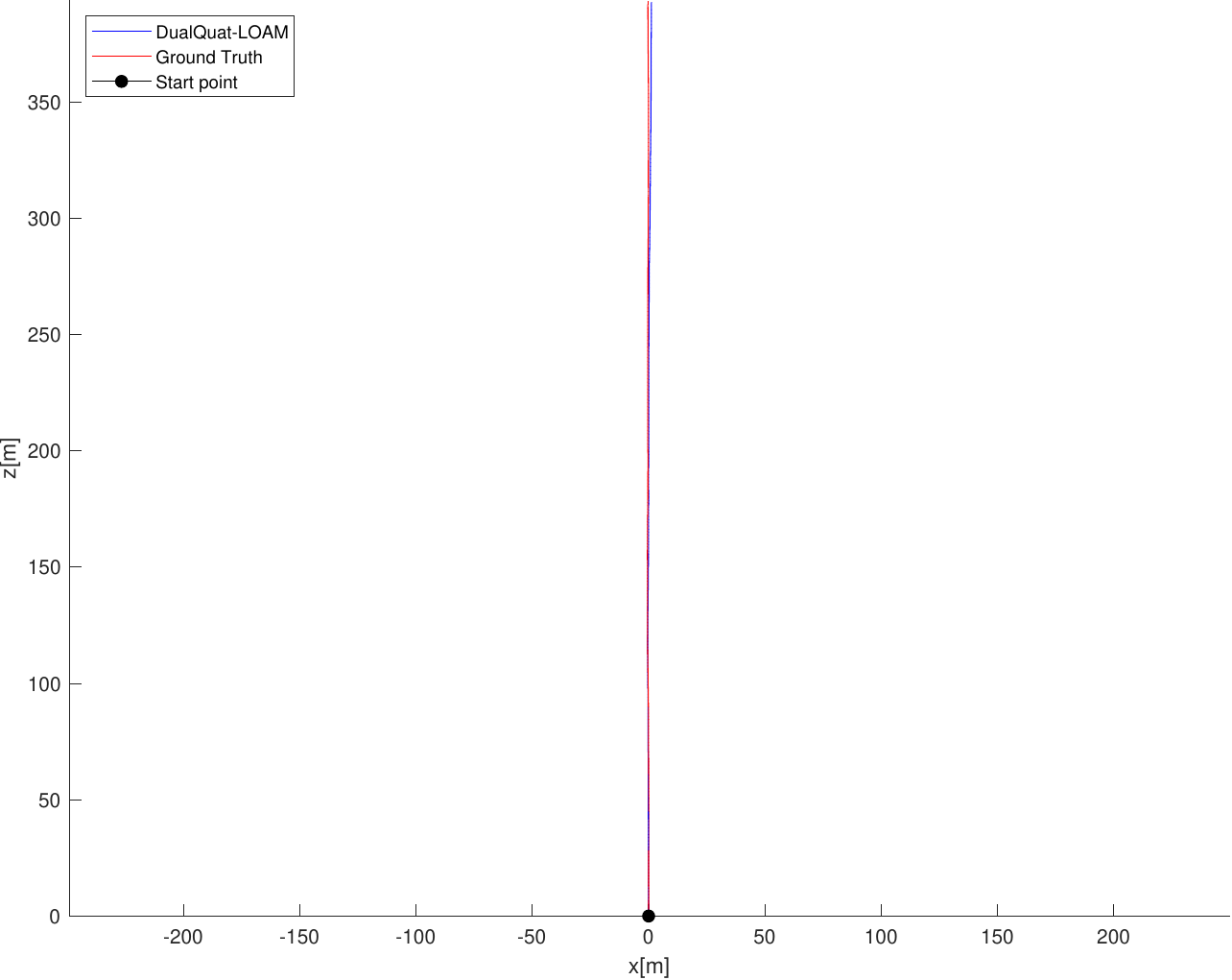}
         \label{fig:fig_kitti_04}
     \end{subfigure}
     \begin{subfigure}[b]{\imgX\textwidth}
         \caption{05}
         \includegraphics[width=\textwidth]{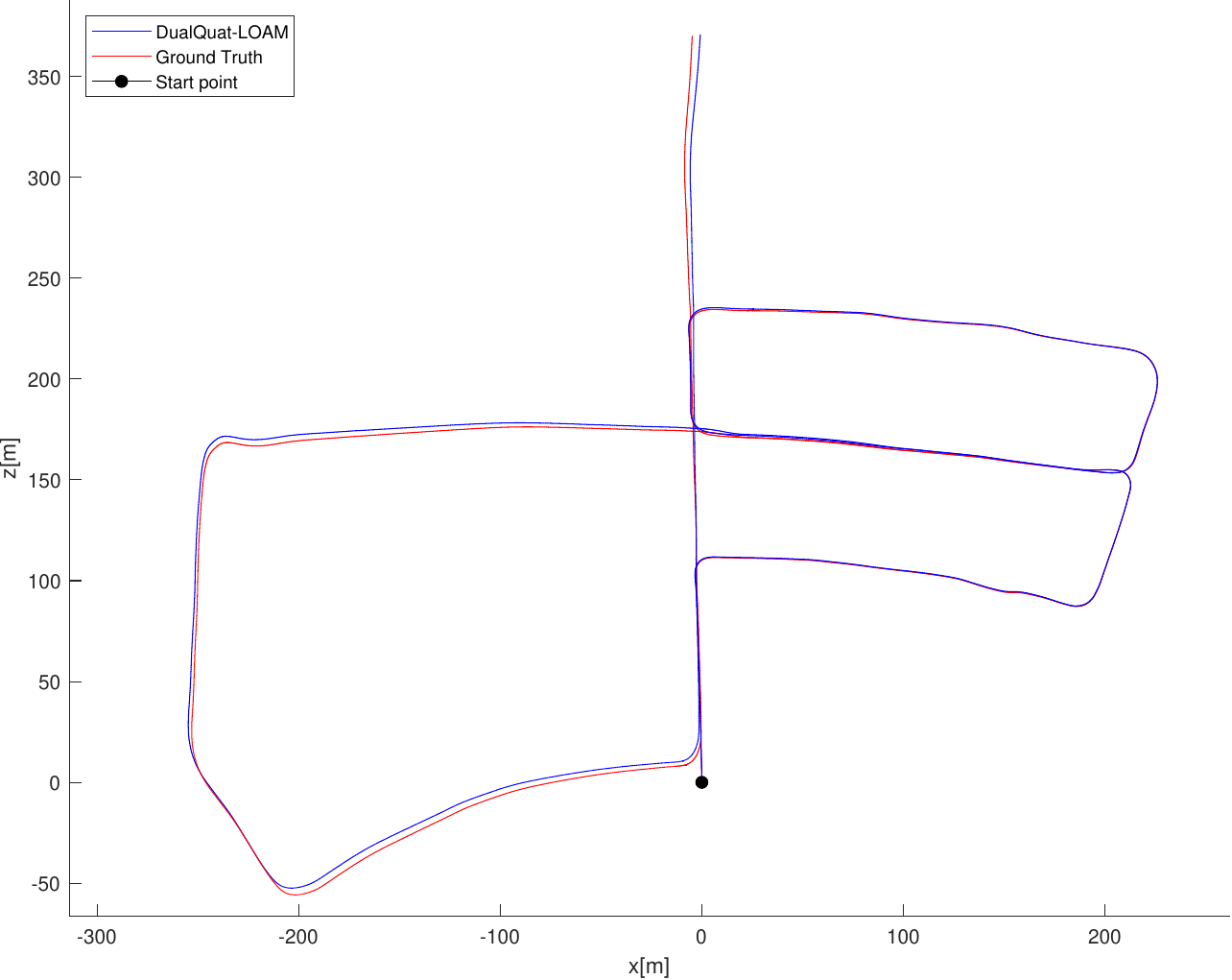}
         \label{fig:fig_kitti_05}
     \end{subfigure}
     
     \begin{subfigure}[b]{\imgX\textwidth}
        \caption{06}
        \includegraphics[width=\textwidth]{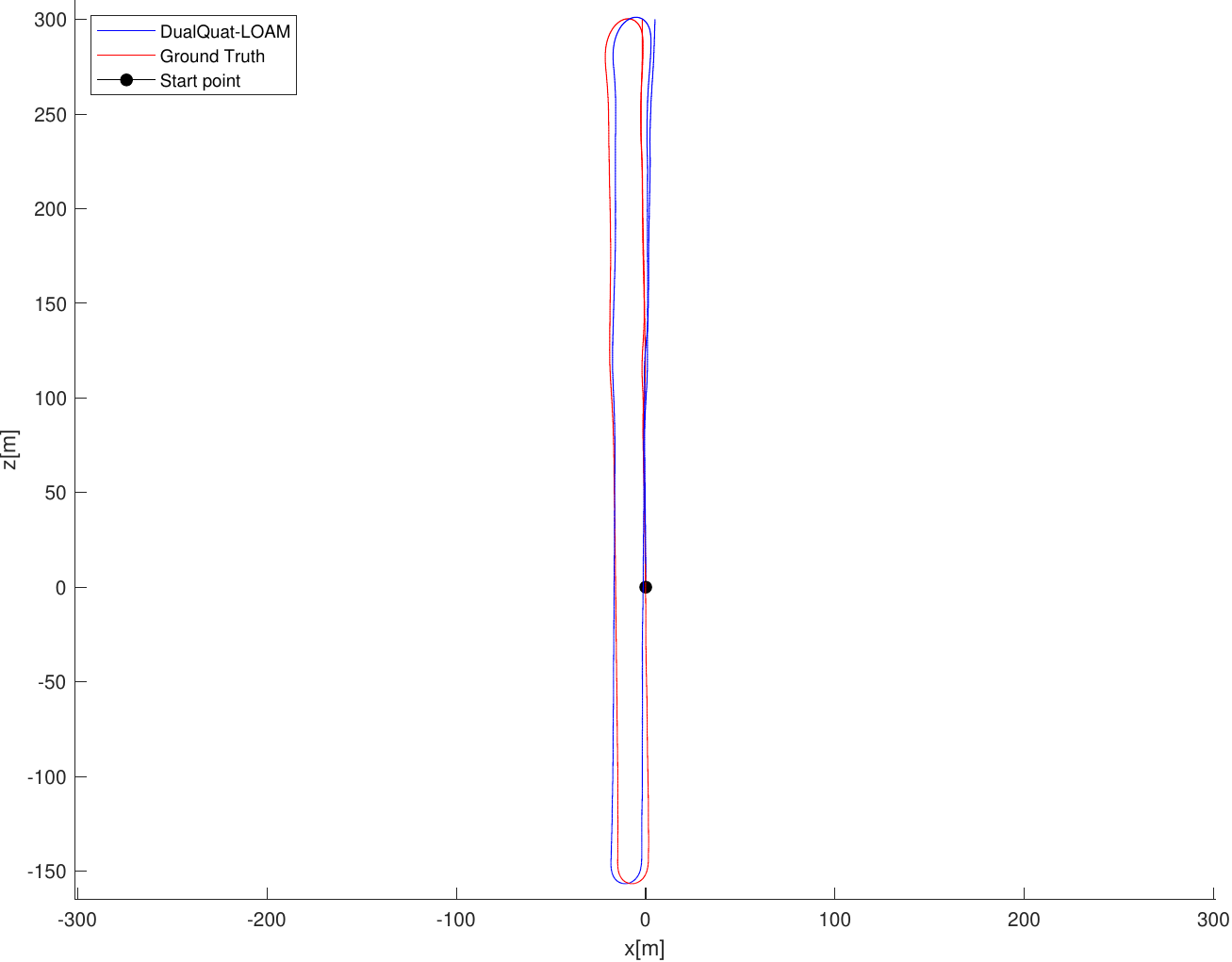}
         \label{fig:fig_kitti_06}
     \end{subfigure}
     \begin{subfigure}[b]{\imgX\textwidth}
         \caption{07}
         \includegraphics[width=\textwidth]{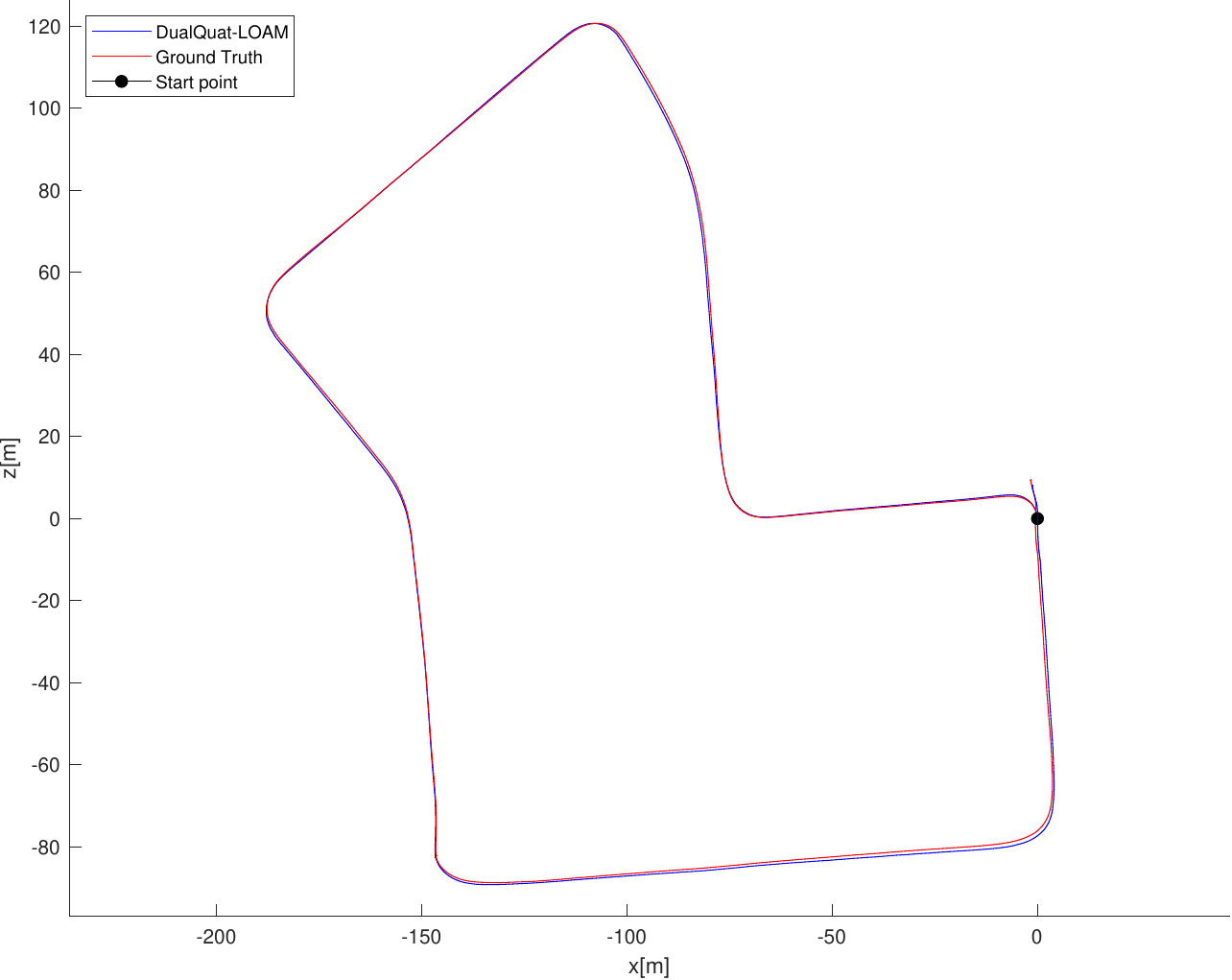}
         \label{fig:fig_kitti_07}
     \end{subfigure}
     \begin{subfigure}[b]{\imgX\textwidth}
         \caption{08}
         \includegraphics[width=\textwidth]{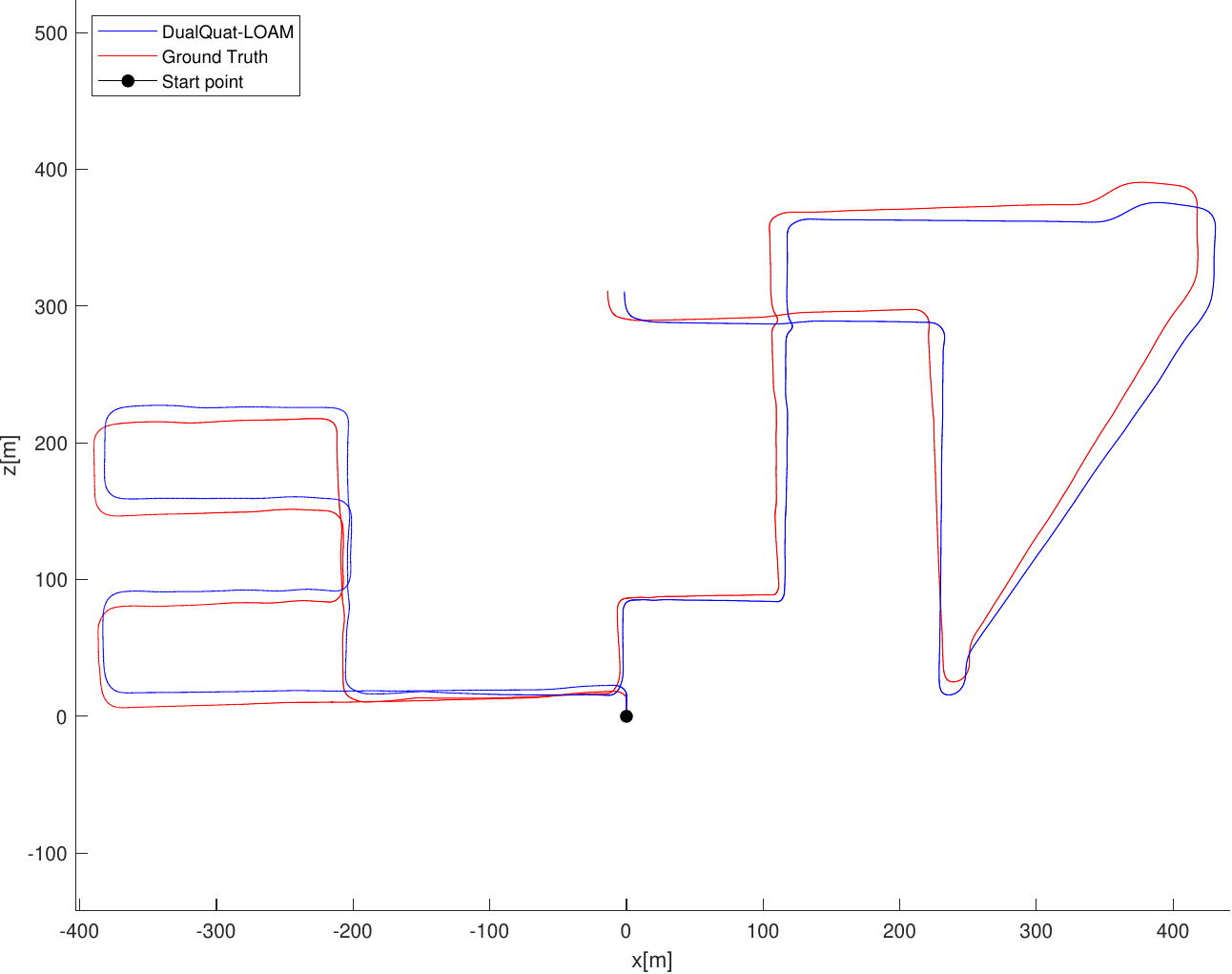}
         \label{fig:fig_kitti_08}
     \end{subfigure}
     \begin{subfigure}[b]{\imgX\textwidth}
         \caption{09}
         \includegraphics[width=\textwidth]{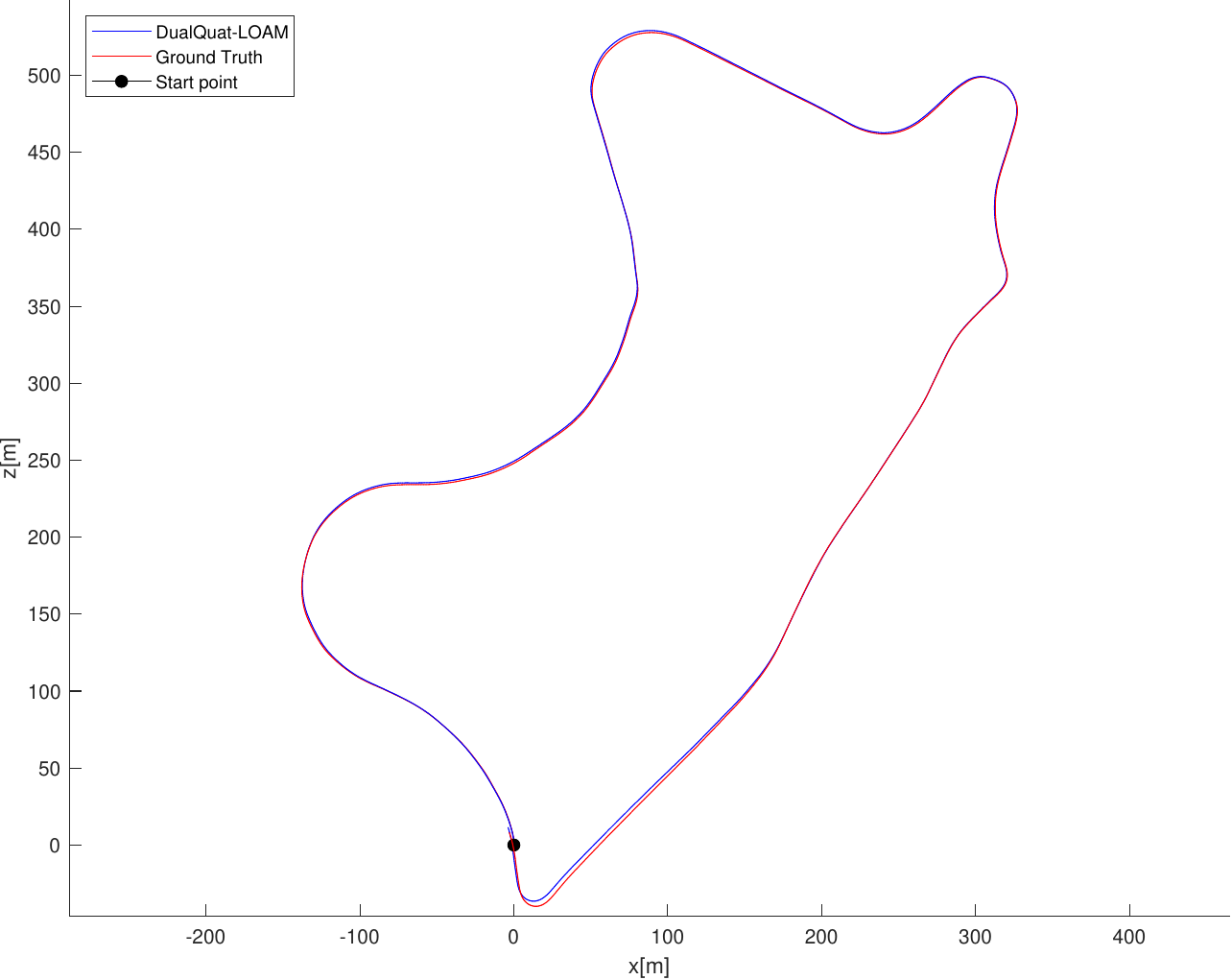}
         \label{fig:fig_kitti_09}
     \end{subfigure}
     \begin{subfigure}[b]{\imgX\textwidth}
         \caption{10}
         \includegraphics[width=\textwidth]{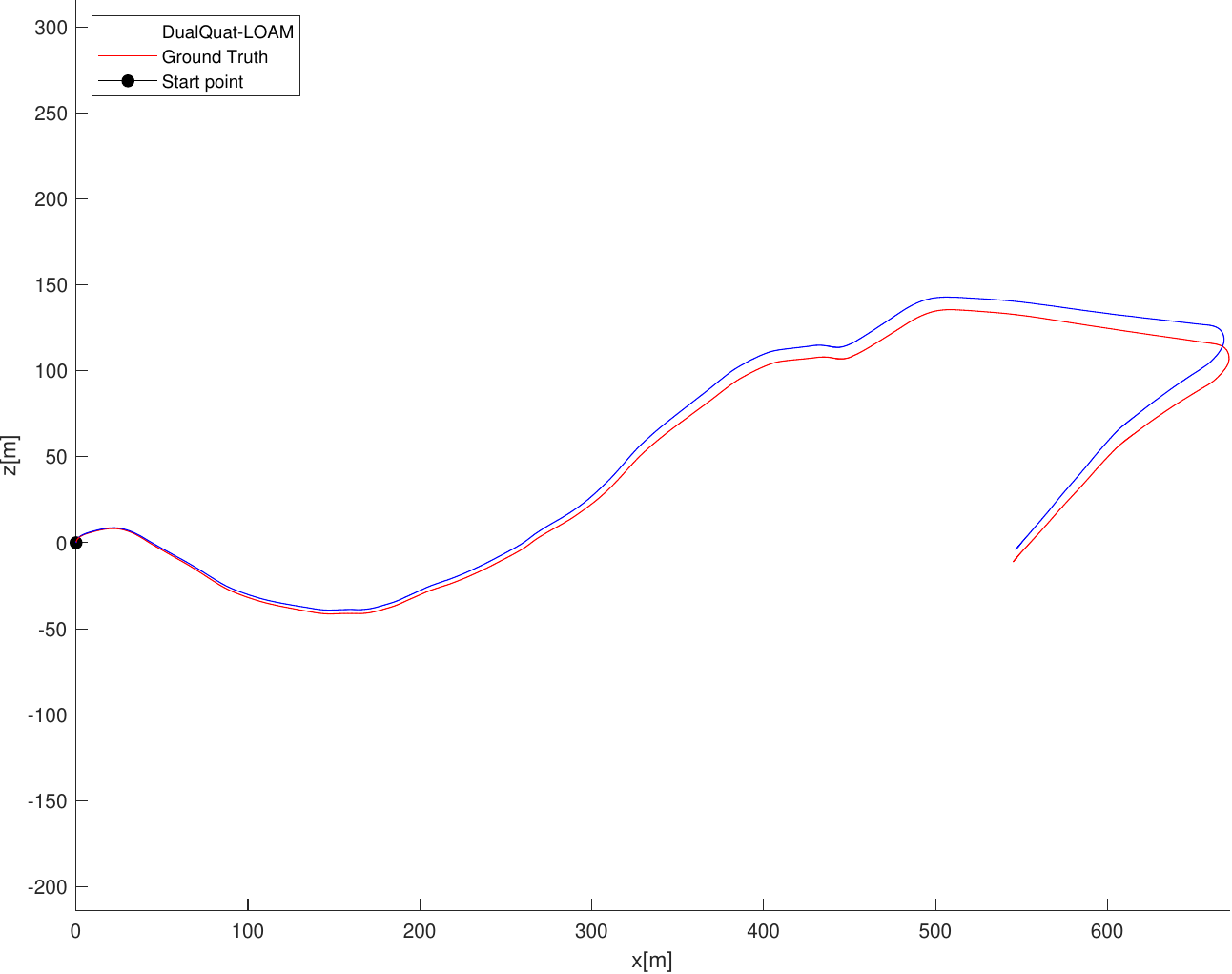}
         \label{fig:fig_kitti_10}
     \end{subfigure}
    \begin{subfigure}[c]{0.4\textwidth}
    \includegraphics[width=\textwidth]{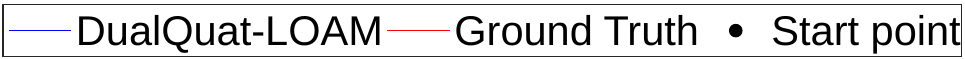}
     \end{subfigure}
\caption{Result of the proposed method on the sequences of KITTI dataset (01-10). The blue line is the estimated odometry with DualQuat-LOAM and the red line is the ground truth.}
\label{fig:fig_kitti_00-10}
\end{figure} 

Our approach was compared to several state-of-the-art methods that use only point clouds for pose estimation, without integrating data from other sensors. The ATE (Absolute Trajectory Error) and ARE (Absolute Rotation Error) values derived from these experiments are displayed in Table \ref{tab:dualquat_loam_comparativa}, which identifies the specific experiments used to obtain the error results. In this table, the comparative results of the 11 sequences of the KITTI dataset are presented. Our DualQuat-LOAM method obtains the lowest translation error (ATE), with a value of $0.79 \%$, compared to other state-of-the-art methods that estimate pose only with point clouds. Furthermore, in terms of rotational error, DualQuat-LOAM ranks third, with a value of $0.39[\text{°}/100~m]$, just $0.03[\text{°}/100~m]$ more than the LiLO method, which ranks second; and $0.09[\text{°}/100~m]$ from the first place which is LIODOM. Fig. \ref{fig:fig_kitti_00-10} shows the results of our DualQuat-LOAM method for each sequence of the KITTI dataset using the edge features, surfaces and STD descriptors; all of them parameterized as dual quaternions, including the optimizer developed with the Ceres library.

Table \ref{tab:dualquat_time} shows the computation times of the DualQuat-LOAM method for each pose estimation stage. The computation time has been recorded for each pose estimation of the 11 sequences in the KITTI dataset, resulting in an average run time of $53[ms]$. This makes DualQuat-LOAM a real-time pose estimation method for LiDAR sensors with data acquisition and update rates of 10 Hz. This experiments can be found in this video\footnote{\href{https://youtu.be/4RgnAGatIVw?feature=shared}{https://youtu.be/4RgnAGatIVw?feature=shared}}.

\begin{table}[ht]
\centering
\footnotesize	
\caption[Average Time for Each Process of the DualQuat-LOAM Method]{Average Time for Each Process of the DualQuat-LOAM Method}
\label{tab:dualquat_time}
\begin{tabular}{@{}lS[table-format=2.2]@{}}
\toprule
Process Stage                         & {Average Time ($[ms]$)} \\ \midrule
Point Cloud Preprocessing             & 3.65  \\
Descriptor Calculation and Matching   & 47.37 \\
Ceres Optimizer                       & 2.15  \\
\textbf{Average Time}                 & \textbf{53.17} \\
\bottomrule
\end{tabular}
\end{table}

\subsection{Evaluation with our research platform }
\begin{figure}[ht]
    \centering
    \includegraphics[width=1.0\linewidth]{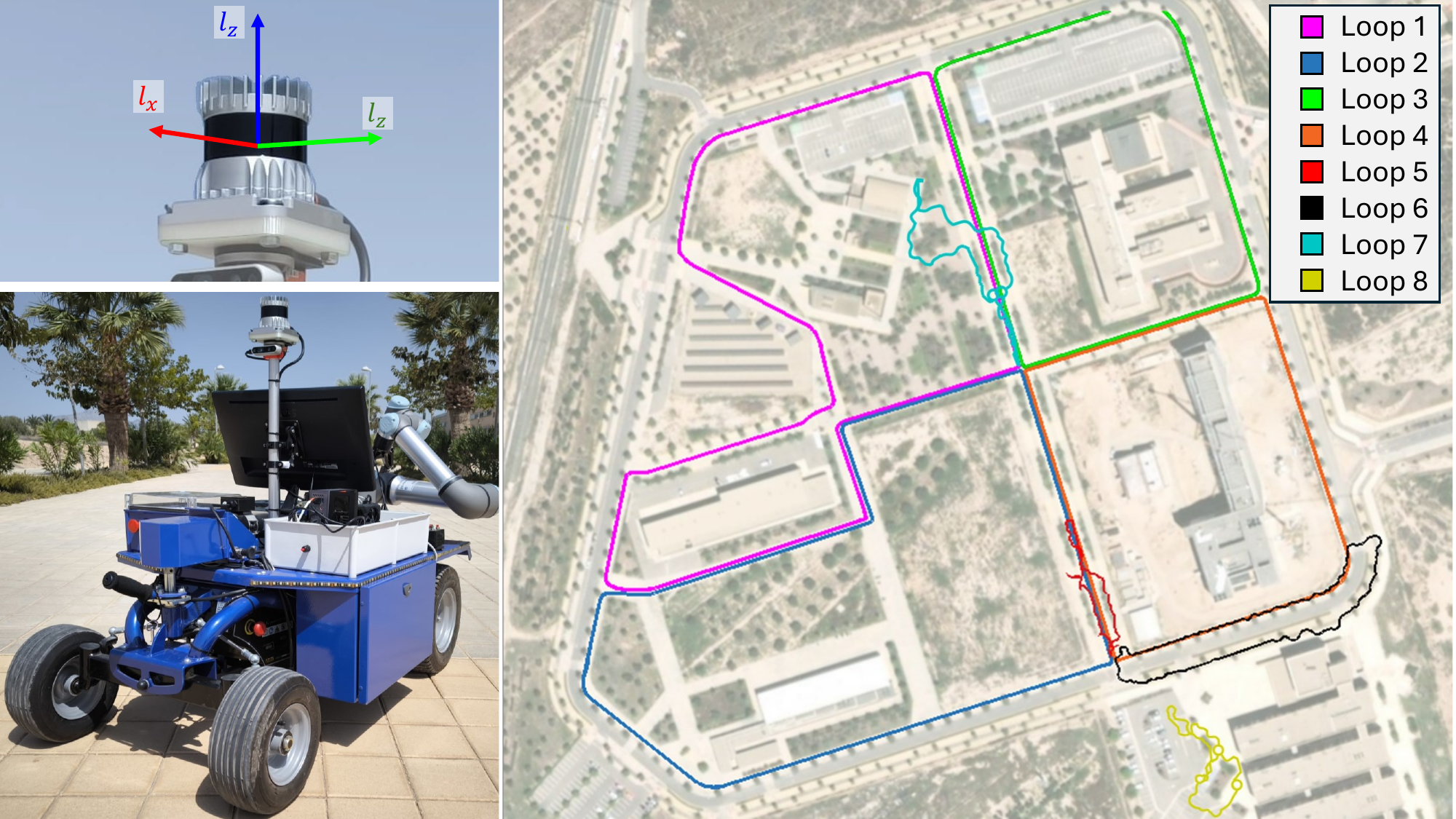}   
    \caption{(\textbf{left}) Ouster OS1-128 LiDAR sensor and roBot for Localization in Unstructured Environments BLUE \cite{BLUE2020deeper}.(\textbf{right}) Path generated by the Muilti-GNSS system from the loops in the area of the Science Park of the University of Alicante.} 
    \label{fig:truth circuits Gnss}
\end{figure}

We also evaluated our approach DualQuat-LOAM presented in this paper on our research platform \textit{BLUE: roBot for Localization in Unstructured Environments} detailed in \cite{BLUE2020deeper} (see Fig. \ref{fig:truth circuits Gnss} on the left). 

This robot has a maximum speed of about $5.0~[km/h]$ and has an Ackermann configuration. It includes an Ouster OS1-128 LiDAR Sensor. For this configuration, the robot is equipped with an Ouster OS1-128 3D LiDAR that we use for the pose estimation of our method, and it also has a Multi-GNSS system with three Ublox Neo-M8N modules to know the routes along which the robot has circulated as shown in Fig. \ref{fig:truth circuits Gnss} on the right. As the BLUE robot speed is considerably lower compared to the KITTI dataset sequences, we set the STD map data window to $h = 4$, which allows us to generate an STD map with up to 480 elements. We compared our LiDAR odometry system, DualQuat-LOAM, with the \mbox{F-LOAM} \cite{wang2021floam}, LiLO \cite{velasco2023lilo}, Fast-LIO2 \cite{xu2022fast} and DLIO \cite{chen2023direct} methods. In addition, we evaluated the performance of our method in pose estimation, both with and without STD descriptors.

The experiments were performed in an unstructured outdoor environment at the Science Park of the University of Alicante, where we evaluated the error in loop closure (see Fig. \ref{fig:truth circuits Gnss} on the right),  generating eight closed loops covering a total distance of approximately $3.6~[km]$. Due to the fact that the experimental environment consisted of closed-loop trajectories and we did not have ground truth data for position evaluation, the analysis focused on the cumulative drift of the different pose estimation methods. To evaluate performance, a loop closure was implemented based on visual markers placed on the ground, which allowed verifying that the robot returned to the same position at the beginning and end of each loop. In this way, the odometry error was evaluated by measuring the deviation from the initial and final position.

\begin{figure*}[b]
\centering
\newcommand\imgVlp{0.22}
\captionsetup[subfigure]{justification=centering}
     \begin{subfigure}[c]{\imgVlp\textwidth}
         \caption{Loop 1}
         \includegraphics[width=\textwidth]{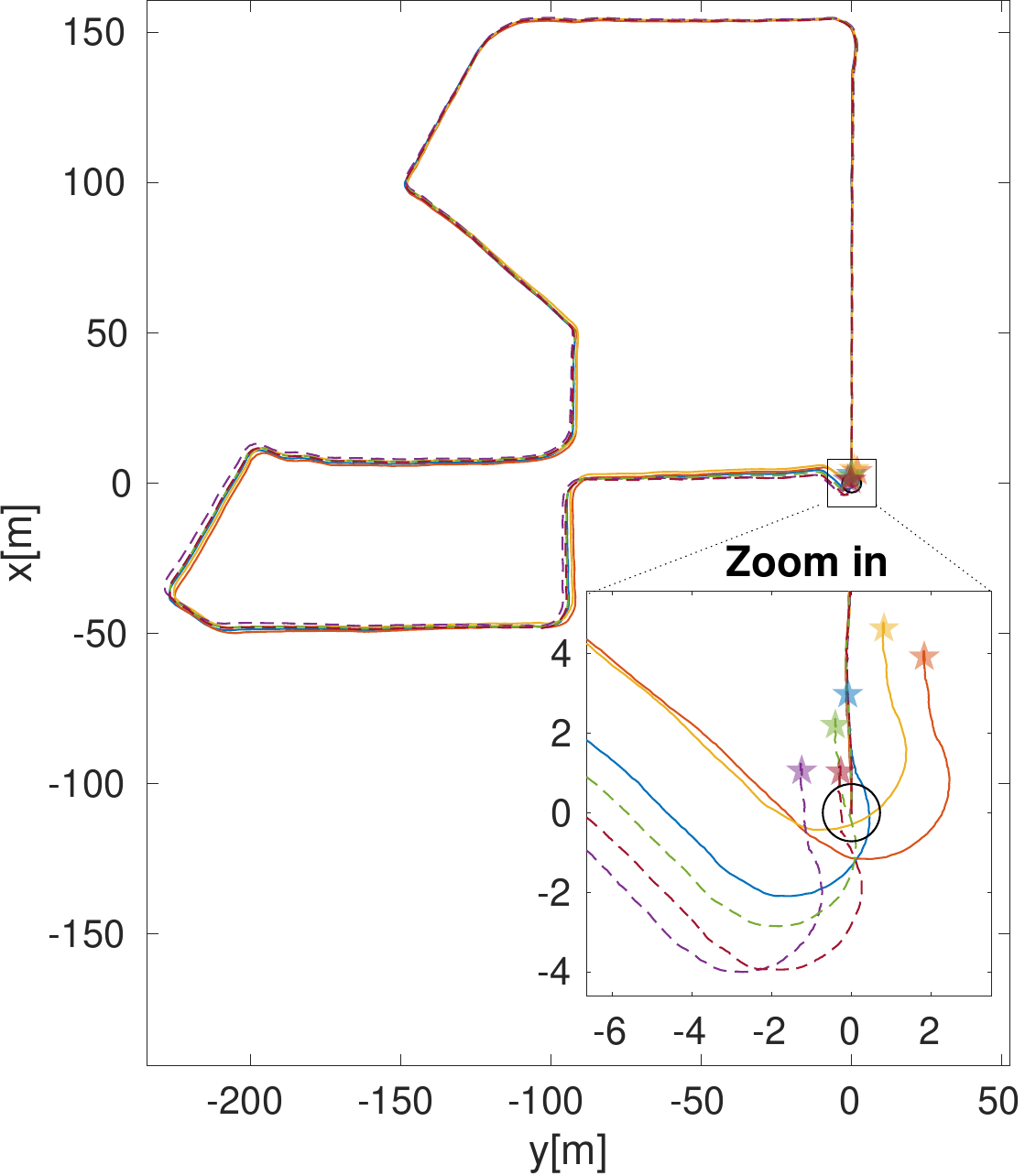}
         \label{fig:fig_Loop_01}
     \end{subfigure}
     \begin{subfigure}[c]{\imgVlp\textwidth}
         \caption{Loop 2}
         \includegraphics[width=\textwidth]{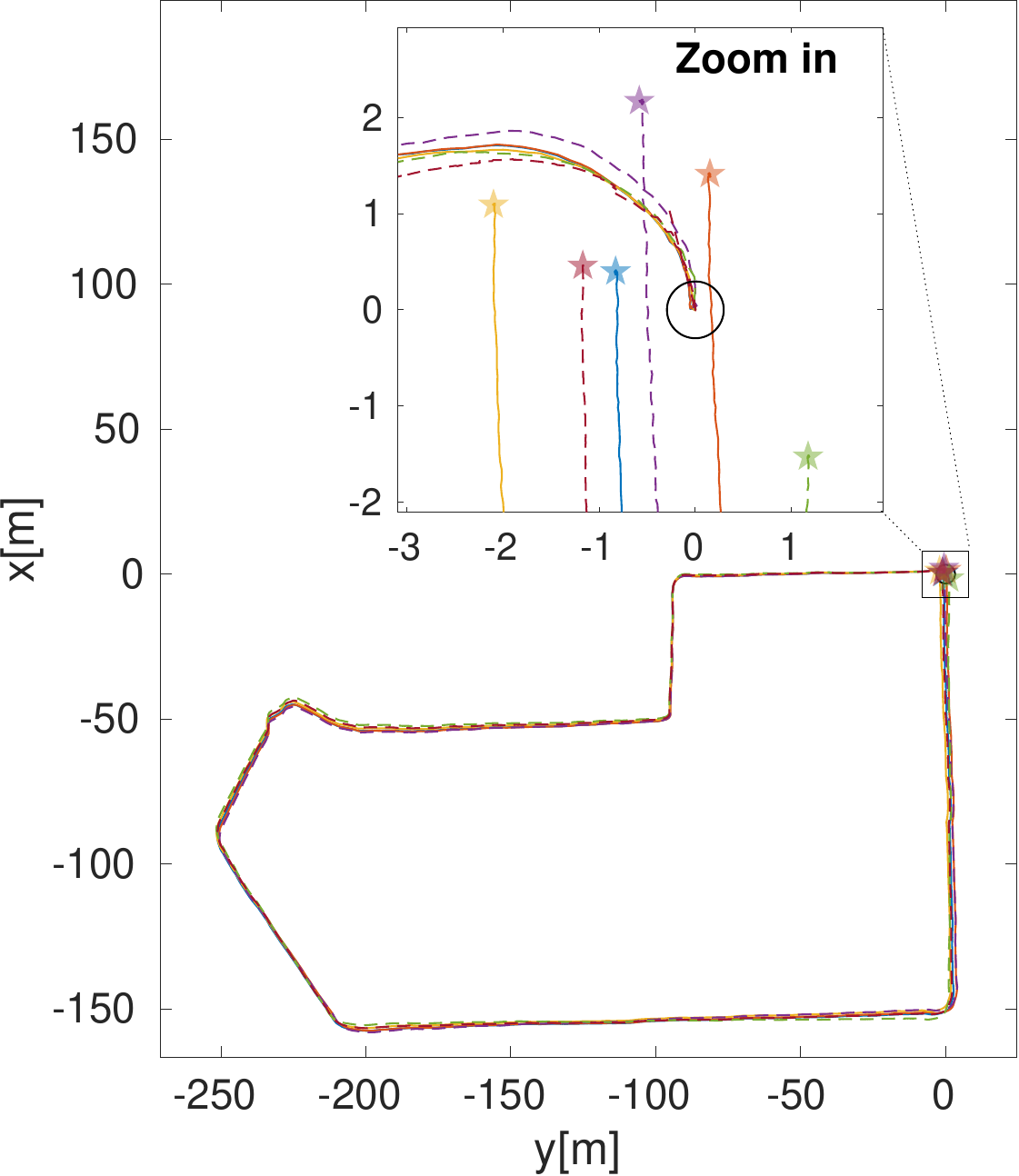}
         \label{fig:fig_Loop_02}
 \end{subfigure}
     \begin{subfigure}[c]{\imgVlp\textwidth}
         \caption{Loop 3}
         \includegraphics[width=\textwidth]{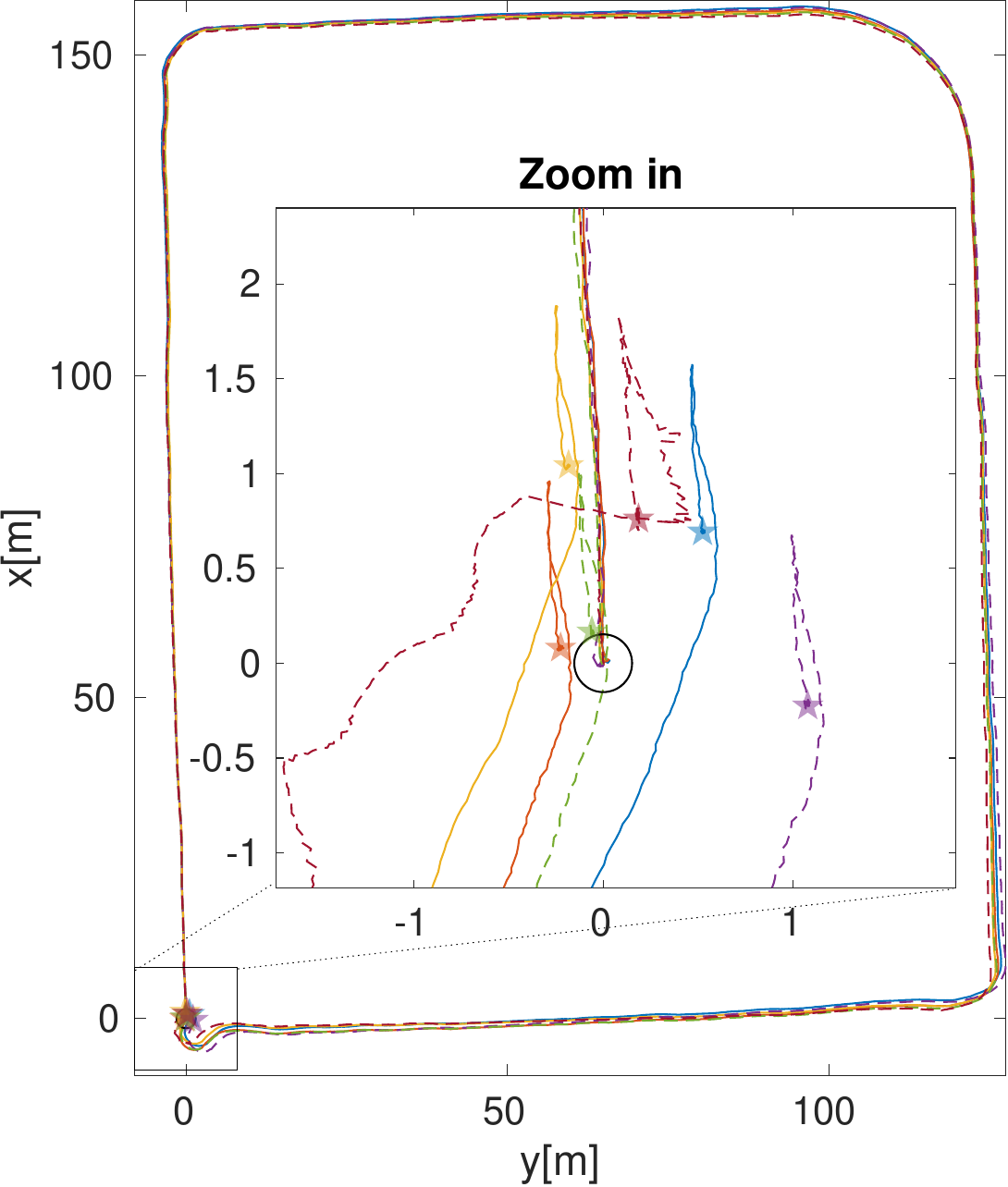}
         \label{fig:fig_Loop_03}
     \end{subfigure}
     \begin{subfigure}[c]{\imgVlp\textwidth}
         \caption{Loop 4}
         \includegraphics[width=\textwidth]{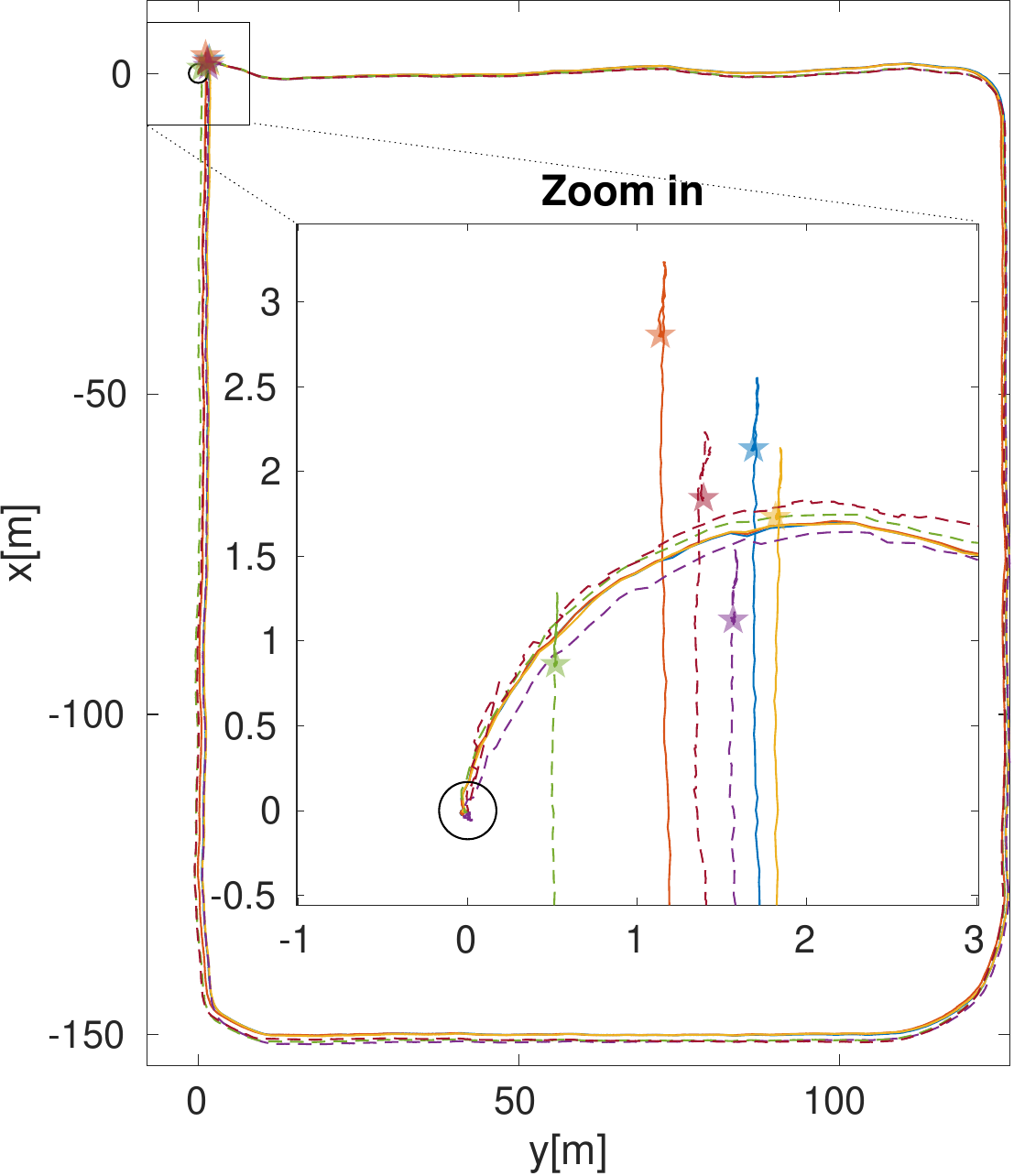}
         \label{fig:fig_Loop_04}
     \end{subfigure}
     \begin{subfigure}[c]{\imgVlp\textwidth}
         \caption{Loop 5}
         \includegraphics[width=\textwidth]{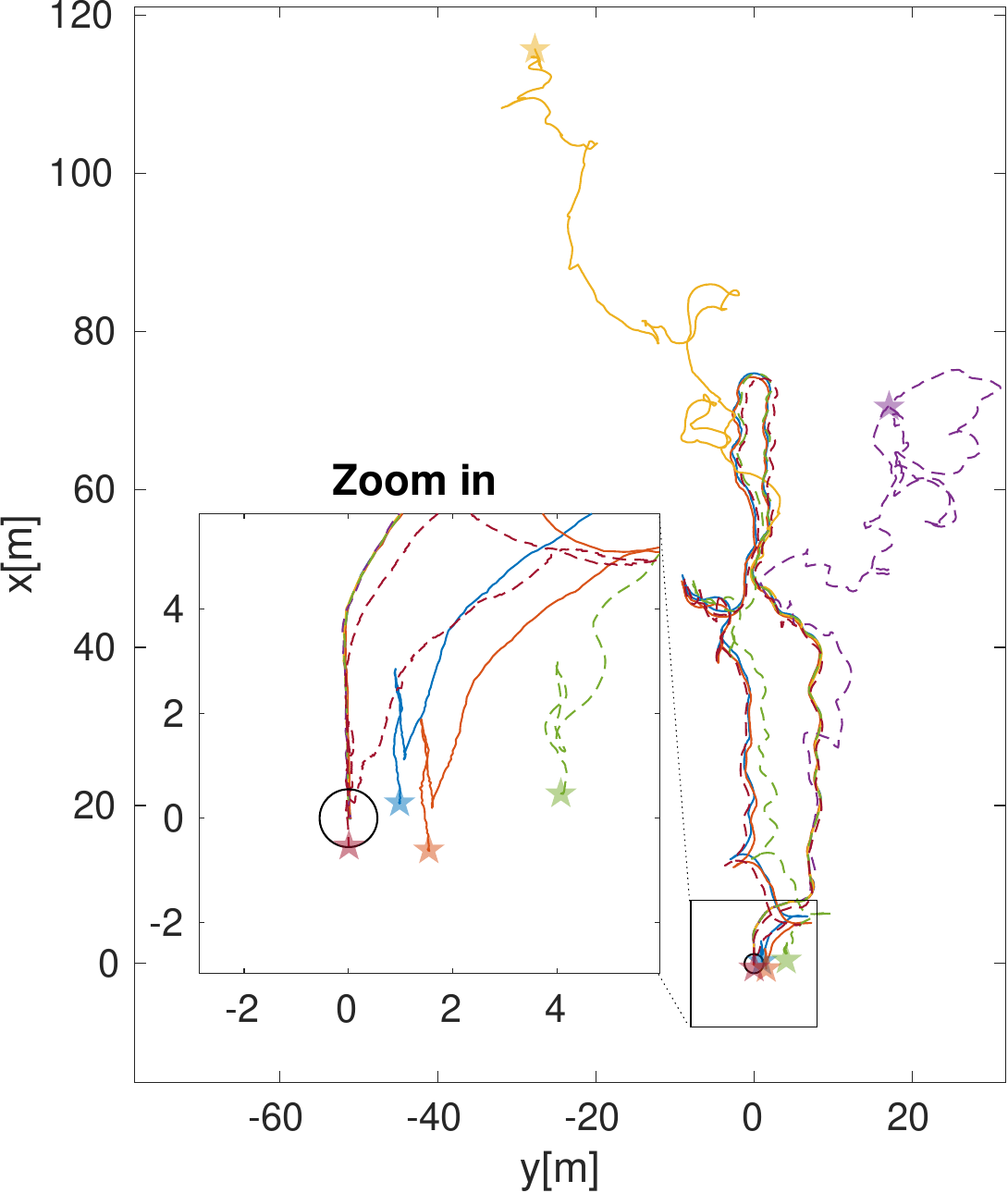}
         \label{fig:fig_Loop_05}
     \end{subfigure}
     \begin{subfigure}[c]{\imgVlp\textwidth}
         \caption{Loop 6}
         \includegraphics[width=\textwidth]{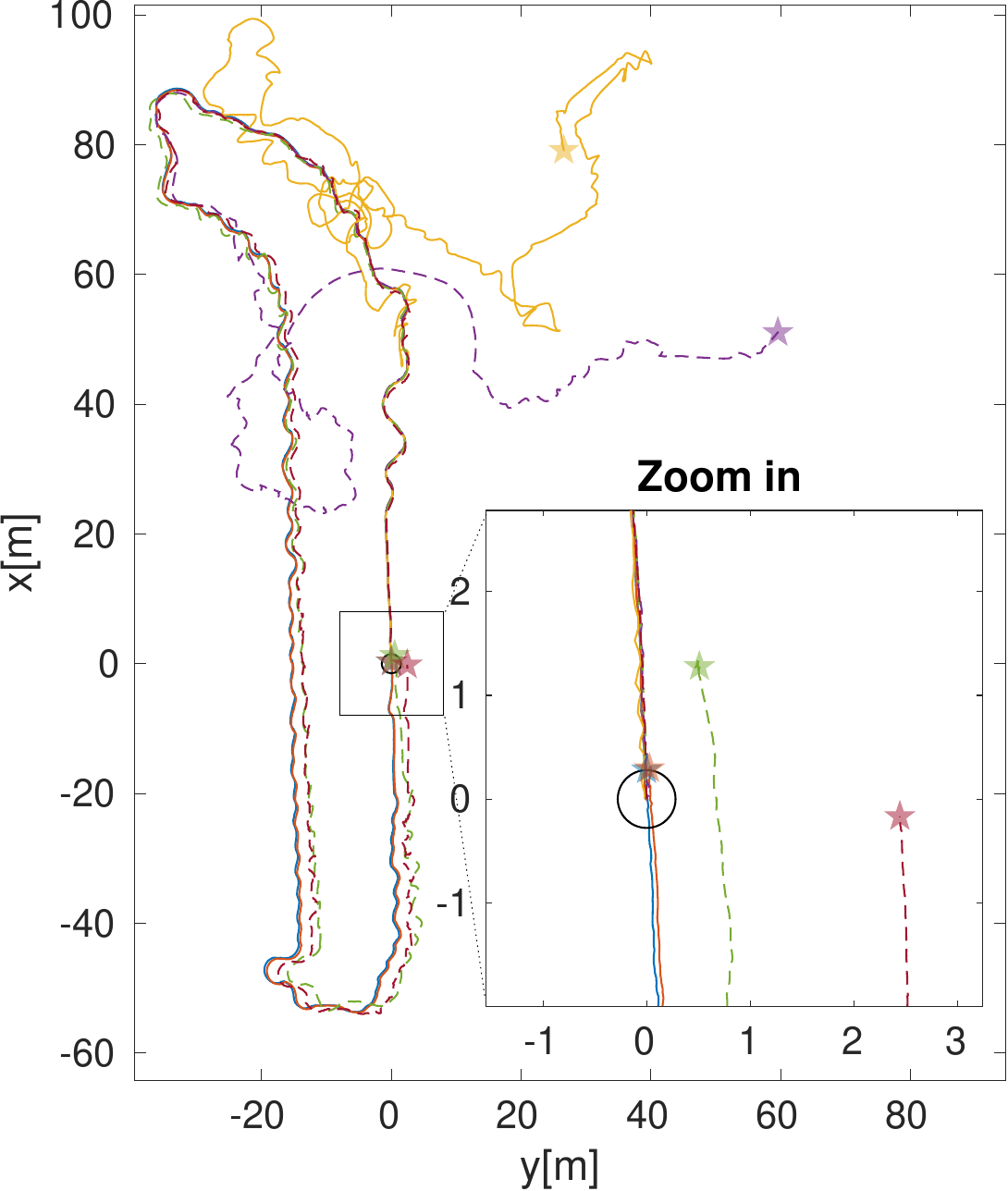}
         \label{fig:fig_Loop_06}
     \end{subfigure}
     \begin{subfigure}[c]{\imgVlp\textwidth}
         \caption{Loop 7}
         \includegraphics[width=\textwidth]{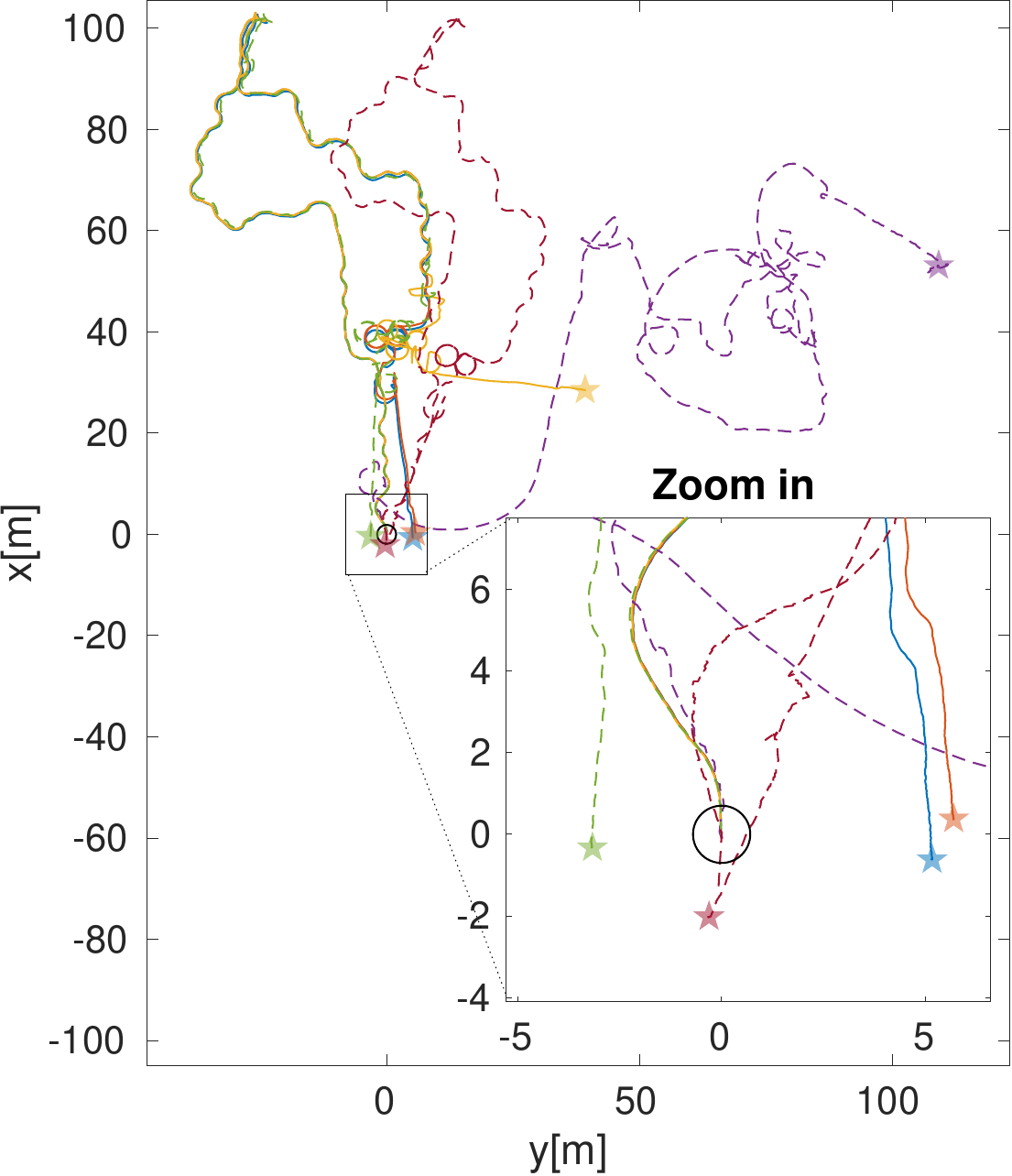}
         \label{fig:fig_Loop_09}
     \end{subfigure}
     \begin{subfigure}[c]{\imgVlp\textwidth}
         \caption{Loop 8}
         \includegraphics[width=\textwidth]{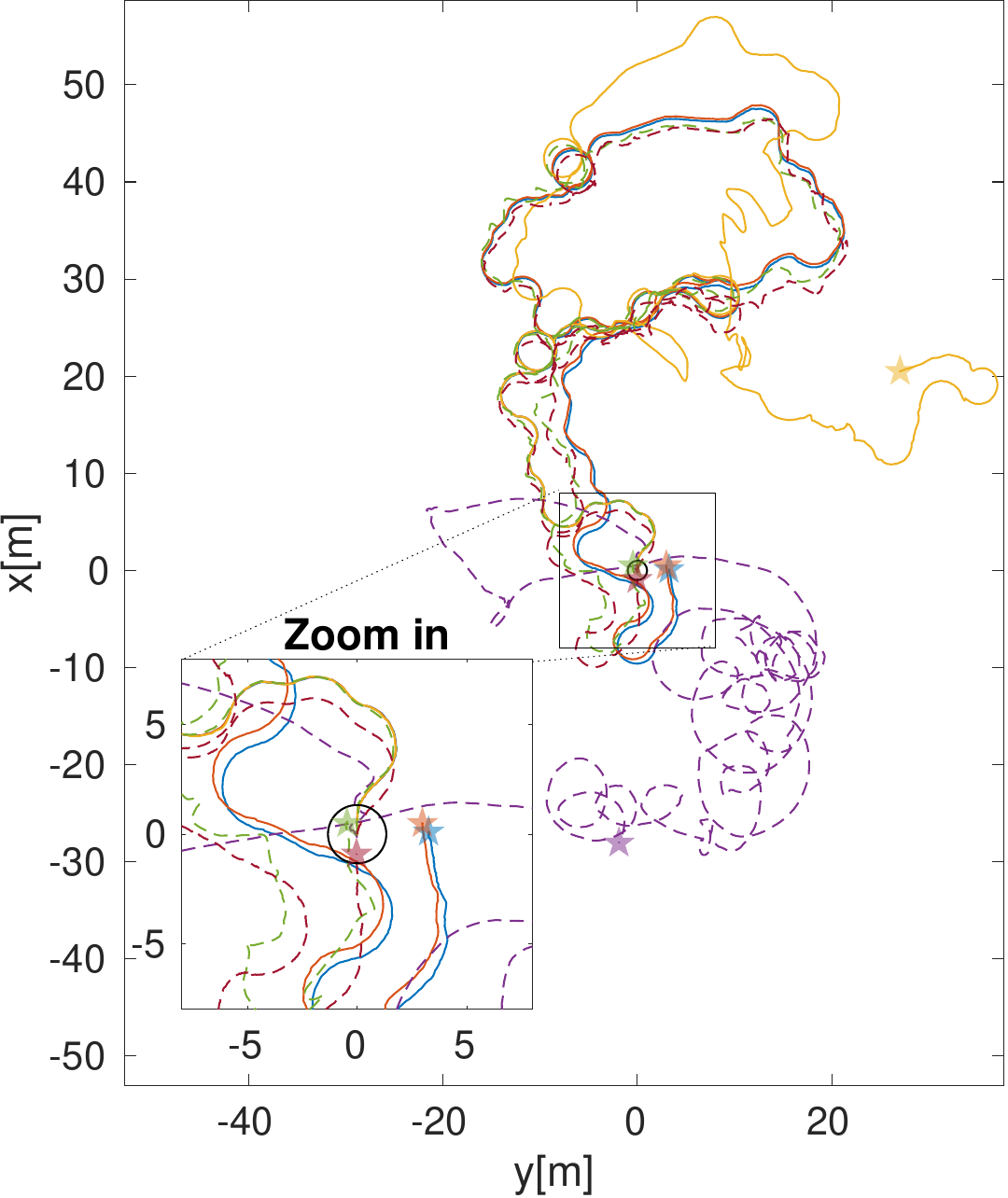}
         \label{fig:fig_Loop_10}
     \end{subfigure}     
     \\
     \begin{subfigure}[c]{0.6\textwidth}
         \includegraphics[width=\textwidth]{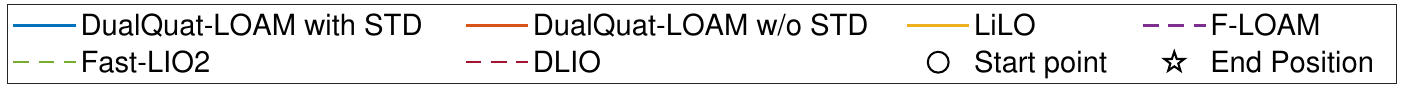}
     \end{subfigure}     
\caption{Benchmark results of our LiDAR odometry method and state-of-the-art methods for pose estimation in our BLUE development platform. All experiments were carried out in closed circuits outside the Science Park of the University of Alicante.}
\label{fig:all_loops}
\end{figure*}

It should be noted that two of the comparative methods, Fast-LIO2 and DLIO , integrate an IMU sensor for pose estimation. These methods are particularly useful to compare our proposal in circuits with several closed curves, since, as shown in the results, methods such as \mbox{F-LOAM} and LiLO present difficulties to estimate the pose in this type of trajectories. In the table \ref{tab:Scientific_Park table} shows the closing errors of the eight loops and Fig. \ref{fig:all_loops} shows the trajectory of each of the LiDAR odometry methods highlighting the closing of the loops. 

\begin{table}
\caption{The value \textbf{d} is the modulus of the initial and final position of each experiment.}
    \centering
    \scriptsize    
    \resizebox{\columnwidth}{!}{%
    \begin{tabular}{cccccccccc}
        \toprule
        \multicolumn{2}{c}{\textbf{Loop}} & \textbf{1} & \textbf{2} & \textbf{3} & \textbf{4} & \textbf{5} & \textbf{6} & \textbf{7} & \textbf{8}\\
        \multicolumn{2}{c}{\textbf{Len. (m)}} & 920 & 770 & 560 & 540 & 209 & 376 & 379 & 284\\ 
        \hline
        & x & -1.25 & -0.58 & 1.08 & 1.57 & 17.09 & 59.58 & 109.06 & -1.90\\
        \textbf{F-LOAM} & y & 1.05 & 2.17 & -0.22 & 1.13 & 70.43 & 51.05 & 53.10 & -28.06\\
        \cite{wang2021floam} & z & -0.55 & -3.15 & -0.85 & -0.78 & -2.81 & -0.23 & -6.30 & -0.36\\
        & \textbf{d} & \textbf{1.72} & \textbf{3.87} & \textbf{1.39} & \textbf{2.08} & \textbf{72.53} & \textbf{78.46} & \textbf{121.46} & \textbf{28.12}\\ 
        \hline
        & x & 0.82 & -2.10 & -0.18 & 1.81 & -27.72 & 26.57 & 39.21 & 27.04\\
        \textbf{LiLO} & y & 4.63 & 1.09 & 1.04 & 1.73 & 115.64 & 79.15 & 28.51 & 20.52\\
        \cite{velasco2023lilo} & z & -3.07 & -0.41 & -0.60 & 1.37 & -4.93 & -0.61 & -0.61 & -0.04\\
        & \textbf{d} & \textbf{5.61} & \textbf{2.40} & \textbf{1.22} & \textbf{2.86} & \textbf{119.02} & \textbf{83.49} & \textbf{48.48} & \textbf{33.94}\\ 
        \hline
        & x & -0.40 & 1.17 & -0.06 & 0.52 & 4.07 & 0.51 & -3.16 & -0.46\\
        \textbf{FAST-LIO2} & y & 2.20 & -1.53 & 0.17 & 0.86 & 0.47 & 1.28 & -0.33 & 0.50\\
        \cite{xu2022fast} & z & -2.49 & -3.10 & -0.01 & -0.03 & -0.03 & -0.03 & -0.04 & -0.06\\
        & \textbf{d} & \textbf{3.35} & \textbf{3.65} & \textbf{0.18} & \textbf{1.01} & \textbf{4.10} & \textbf{1.38} & \textbf{3.18} & \textbf{0.68}\\
        \hline
        & x & -0.27 & -1.17 & 0.19 & 1.39 & 0.01 & 2.44 & -0.31 & -0.03\\
        \textbf{DLIO} & y & 1.03 & 0.46 & 0.76 & 1.84 & -0.53 & -0.17 & -2.01 & -0.91\\
        \cite{chen2023direct} & z & -3.68 & -6.53 & 3.81 & 0.82 & 0.03 & 0.02 & 0.12 & 0.04\\
        & \textbf{d} & \textbf{3.83} & \textbf{6.65} & \textbf{3.89} & \textbf{2.45} & \textbf{0.53} & \textbf{2.45} & \textbf{2.04} & \textbf{0.91}\\
        \hline
        & x & 1.31 & -3.57 & 0.65 & 2.61 & 0.90 & 0.22 & 5.69 & 2.98\\
        \textbf{DualQuat} & y & 6.15 & 4.07 & 0.20 & 1.83 & 0.22 & 0.29 & 0.38 & 0.49\\
        \textbf{LOAM} & z & -5.12 & -3.44 & -0.16 & -1.64 & 0.01 & -0.12 & -0.06 & -0.02\\
        \textbf{w/o STD} & \textbf{d} & \textbf{8.11} & \textbf{6.41} & \textbf{0.70} & \textbf{3.59} & \textbf{0.93} & \textbf{0.38} & \textbf{5.70} & \textbf{3.02}\\
        \hline
        & x & 0.54 & -1.67 & 0.12 & 1.83 & 0.32 & 0.19 & 5.16 & 3.26\\
        \textbf{DualQuat} & y & 2.64 & 1.19 & 0.19 & 2.47 & -0.39 & 0.01 & -0.62 & 0.10\\
        \textbf{LOAM} & z & -3.10 & 0.66 & -0.36 & -1.38 & -0.06 & -0.38 & -0.15 & -0.05\\
        \textbf{with STD} & \textbf{d} & \textbf{4.11} & \textbf{2.16} & \textbf{0.42} & \textbf{3.37} & \textbf{0.51} & \textbf{0.42} & \textbf{5.20} & \textbf{3.26}\\
        \hline
    \end{tabular}
    }
\label{tab:Scientific_Park table}
\end{table}

In Loops 5, 6, 7 and 8, which contain sharp curves and aggressive movements with high angular displacement, both \mbox{F-LOAM} and LiLO exhibit erratic performance, with loop closure errors of $72.53~[m]$, $78.46~[m]$, $121.46~[m]$ and $28.12~[m]$ for \mbox{F-LOAM}, and $119.02~[m]$, $83.49~[m]$, $48.48~[m]$ and $33.94~[m]$ for LiLO. In contrast, DualQuat-LOAM, which also relies solely on point clouds for pose estimation, shows remarkably superior performance, with loop closure errors of only $0.51~ [m]$ and $0.93~[m]$, $5.20~[m]$ and $3.26~[m]$ on the same circuits. These results underscore the effectiveness of DualQuat-LOAM in complex environments, despite its exclusive reliance on LiDAR.

Furthermore, Table \ref{tab:Scientific_Park table} shows that the DualQuat-LOAM method, when used with STD descriptors, has lower loop closure errors compared to its version without STD descriptors. For example, in Loop 1, the loop closure error is $4.11~[m]$ when STD descriptors are used, compared to $8.11~[m]$ without them. This suggests that STD descriptors provide an improvement by decreasing the drift error, particularly in longer and more complex trajectories, where maintaining spatial coherence is crucial.

\begin{figure}[ht]
    \centering
    \includegraphics[width=1.0\linewidth]{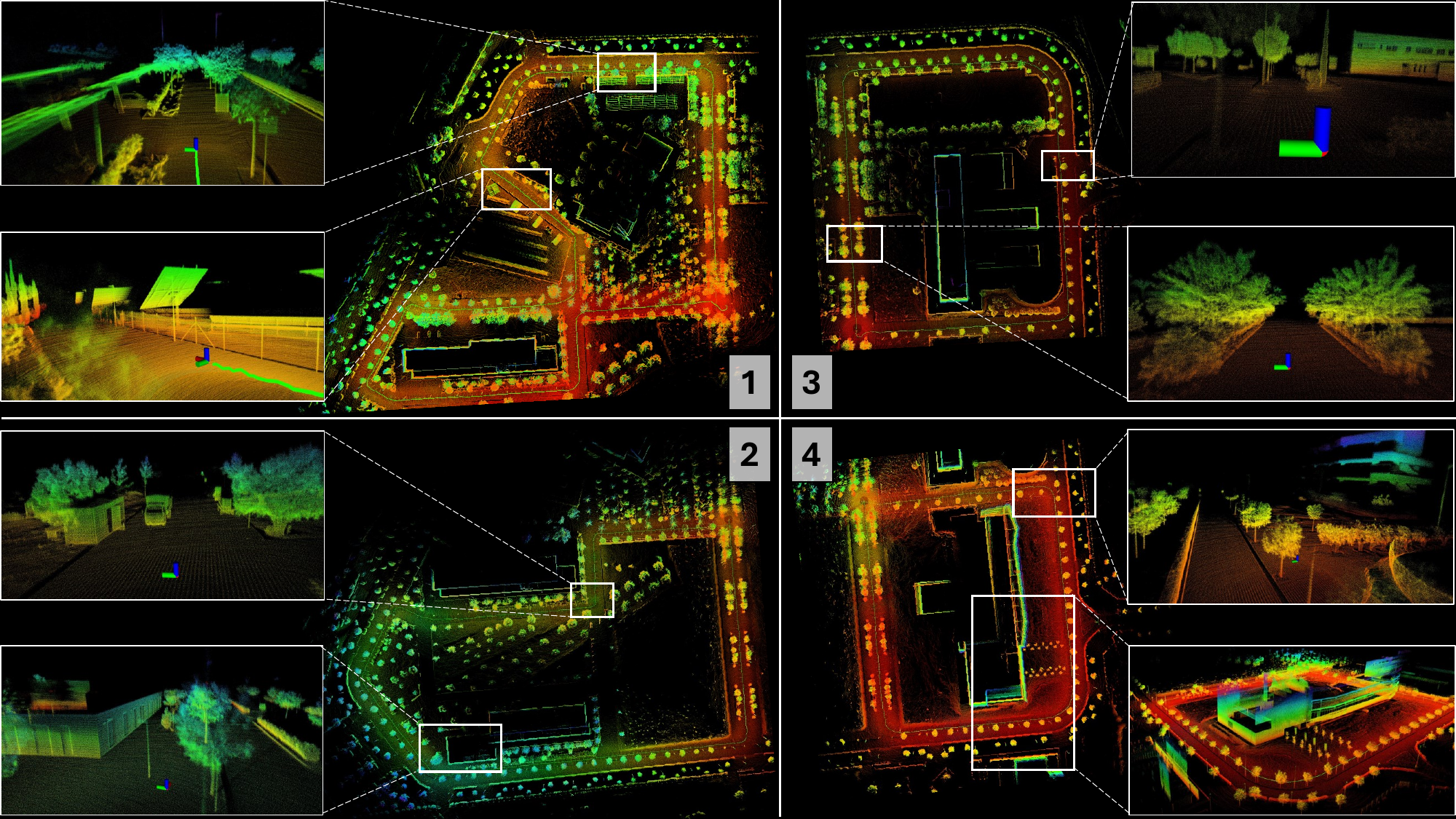}       
    \caption{University of Alicante Science Park. Dense maps generated by our DualQuat-LOAM LiDAR odometry and mapping method. The circuits correspond to loops 1, 2, 3, 4 which were performed in closed loop to estimate the drift error.}    
    \label{fig:dualquat_scientific_park}
\end{figure}

\begin{figure}[ht]
    \centering
    \includegraphics[width=0.85\linewidth]{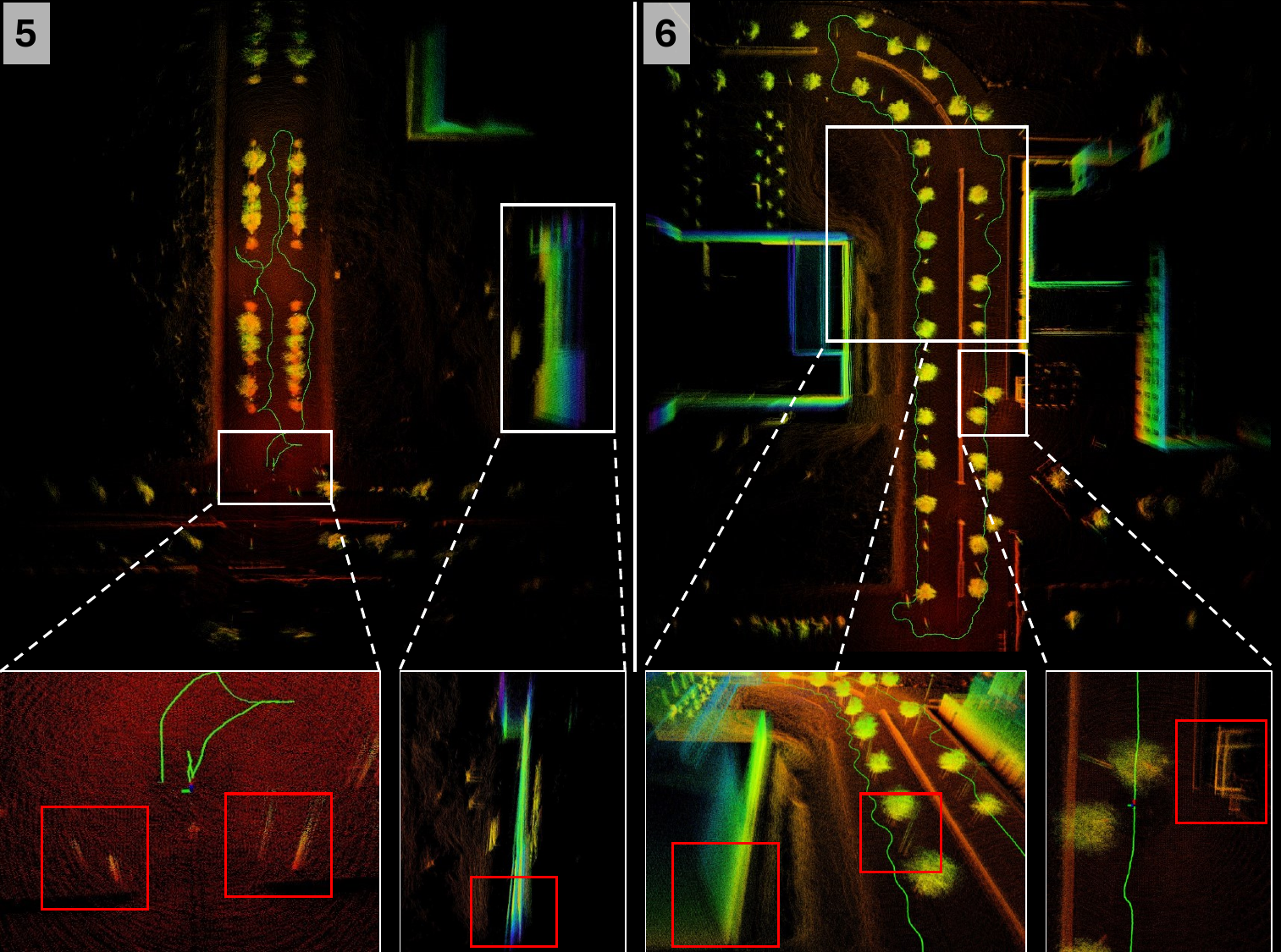}
    \\
    \includegraphics[width=0.85\linewidth]{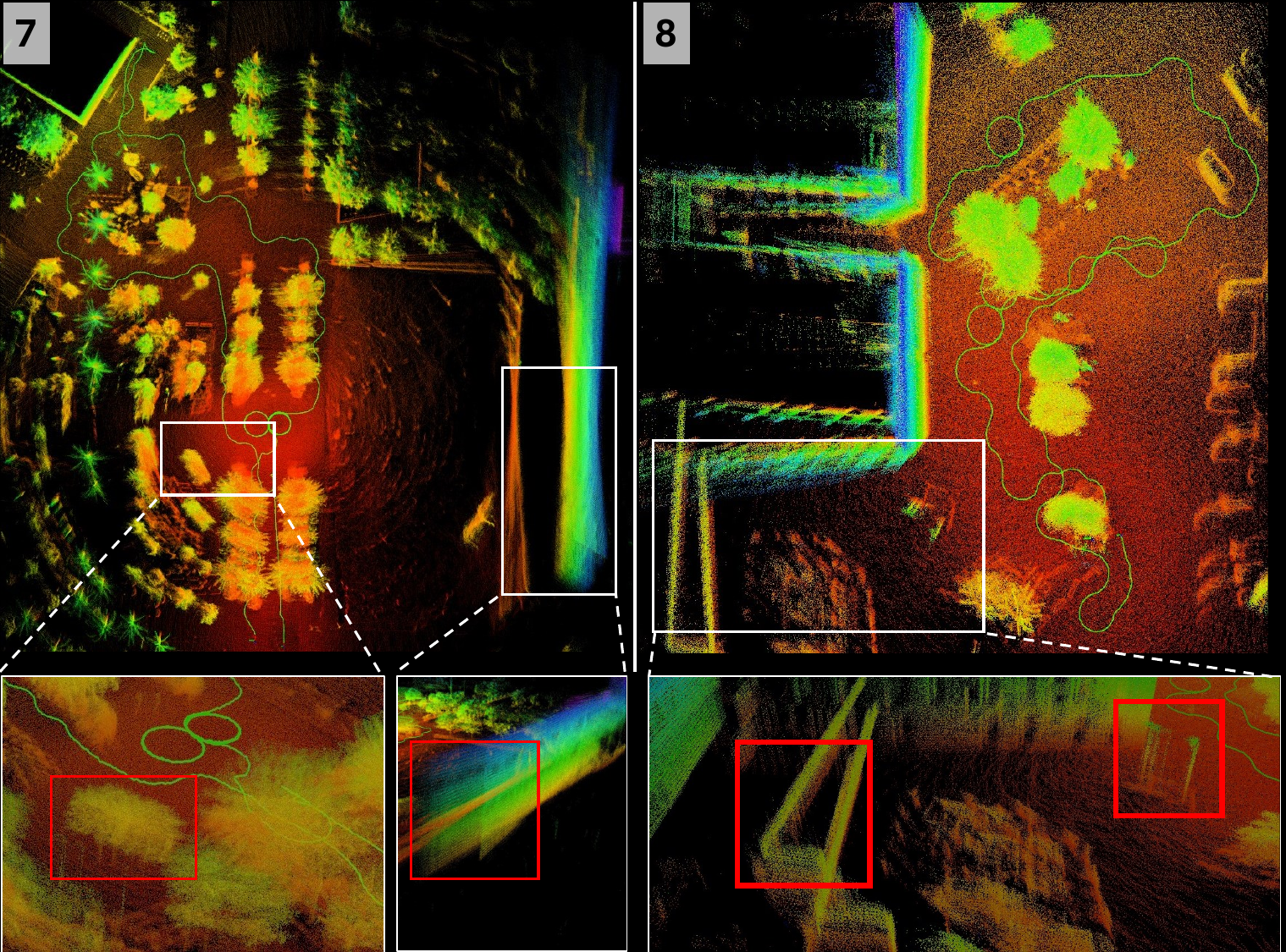}
    \caption{Dense maps generated by our DualQuat-LOAM method for loops 5, 6, 7 and 8, performed with our BLUE development platform. These loops include several turns designed to evaluate the drift error of our method.} 
    \label{fig:dualquat_loop5-8}
\end{figure}

Fast-LIO2 and DLIO, which integrate IMU sensors for pose estimation, show robust performance in most loops, with relatively low loop closure errors. In Loops 5 to 8, which have sharp turns and tight curves, Fast-LIO2 records errors of $4.10~[m]$, $1.38 ~[m]$, $3.18~[m]$ and $0.68~[m]$, while DLIO reaches $0.53 ~[m]$, $2.45 ~[m]$, $2.04~[m]$ and $0.91~[m]$ respectively. However, despite not using IMU and only relying on LiDAR point cloud, DualQuat-LOAM achieves comparable and even superior performance, with loop closure errors of only $0.51~ [m]$ and $0.93~[m]$, $5.20~[m]$ and $3.26~[m]$ on the same circuits. This shows DualQuat-LOAM's ability to maintain low drift error in pose estimation without the need for additional sensors.

Fig. \ref{fig:dualquat_scientific_park} shows the dense maps generated by the DualQuat-LOAM method in the loops 1, 2, 3 and 4. These loops present relatively smooth trajectories, which allows the method to maintain high accuracy in the accumulation of the dense map, as seen in the uniformity and consistency of the data along the routes. In contrast, Fig. \ref{fig:dualquat_loop5-8} presents the dense maps corresponding to loops 5, 6, 7 and 8. These experiments, characterized by sharp curves and aggressive movements with high angular displacement, reveal how these movements generate accumulation errors in the dense map. These errors are visualized in the red boxes at the bottom of the figure, where a distortion in the representation of the environment is observed. 

\subsection{Evaluation with Public Benchmark Datasets}

This section presents the results obtained by evaluating the DualQuat-LOAM method against other odometry methods based on LiDAR-only odometry (LO) and LiDAR-inertial odometry (LIO). The experiments were performed using the datasets: ConSLAM \cite{trzeciak2022conslam} (\texttt{sequence02}, captured with handheld equipment at a construction site), NTU VIRAL \cite{nguyen2022ntu} (\texttt{eee03}, obtained with a drone at a university campus) and HeLiPR \cite{jung2024helipr} (\texttt{Roundabout02}, captured with a car traveling $74471~[m]$  in an urban environment). These experiments were performed following the results and metrics reported in \cite{lee2024lidar}, and are shown in Table \ref{table:benchmark}. Similar to \cite{lee2024lidar}, we use the evo evaluation tool \cite{grupp2017evo} to calculate the ATE metrics of the estimated pose. 

In the ConSLAM dataset, DualQuat-LOAM obtained an ATE error of $5.334 ~[m]$, which is higher than those of DLO ($0.154 ~[m]$) and LeGO-LOAM ($0.263 ~[m]$), indicating that, in this closed environment, DualQuat-LOAM struggles against other methods. However, this value is still notably lower than that of KISS-ICP ($13.517~[m]$), showing that DualQuat-LOAM handles this scenario better than some other LiDAR-only odometry methods. 

In the NTU VIRAL dataset, DualQuat-LOAM presented an ATE error of $1.392~[m]$, which places it behind DLO ($0.142~[m]$) and KISS-ICP ($0.829~[m]$) methods, but surpassing LeGO-LOAM ($8.478~[m]$). However, due to the abrupt movements characteristic of the scenarios captured with UAVs, it is evident that for these cases a LIO method would be necessary to decrease the odometry error. 

\begin{figure*}[ht]
    \centering
    \includegraphics[width=1.0\linewidth]{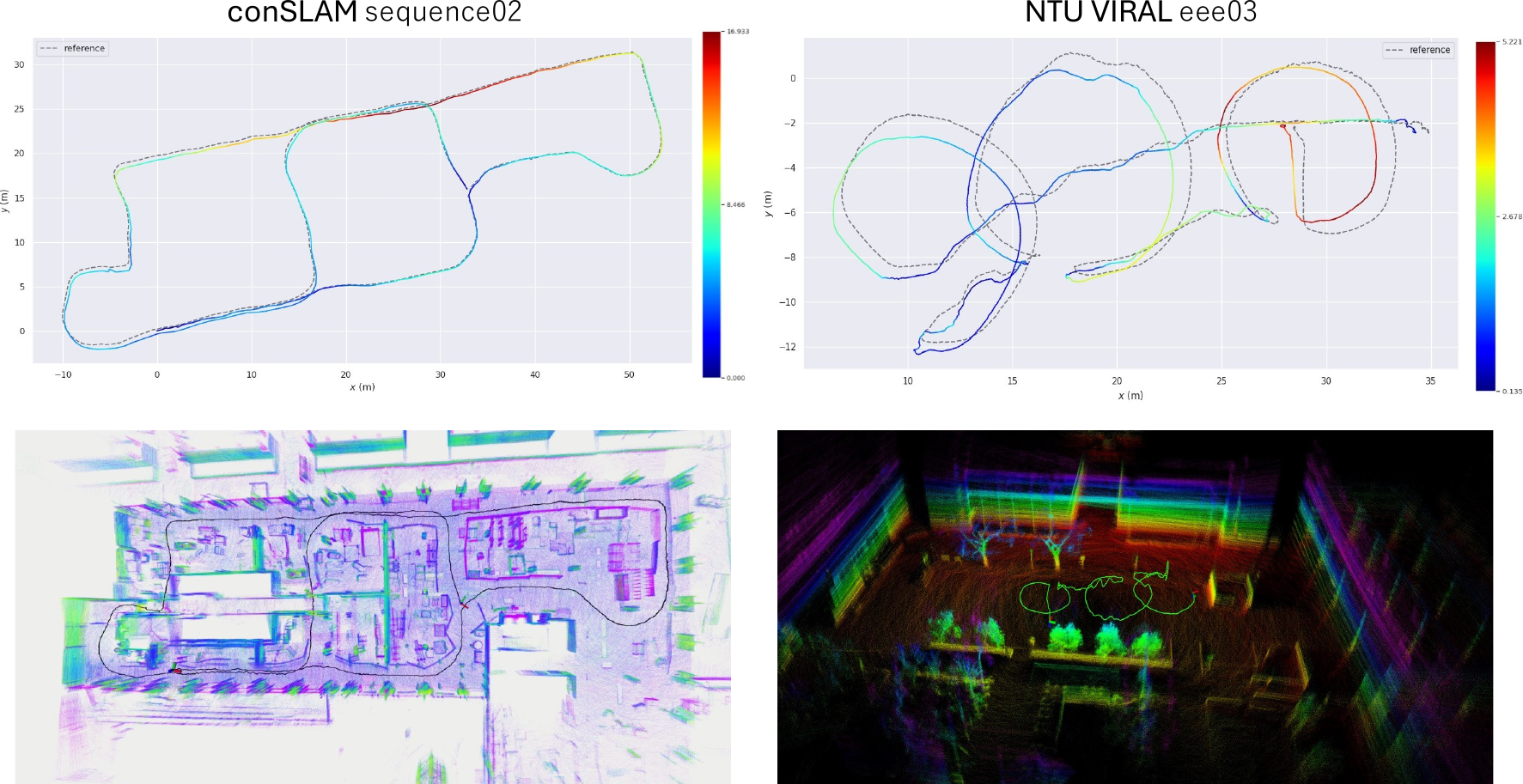}
     \vspace{0.05cm}
     \\
    \includegraphics[width=1.0\linewidth]{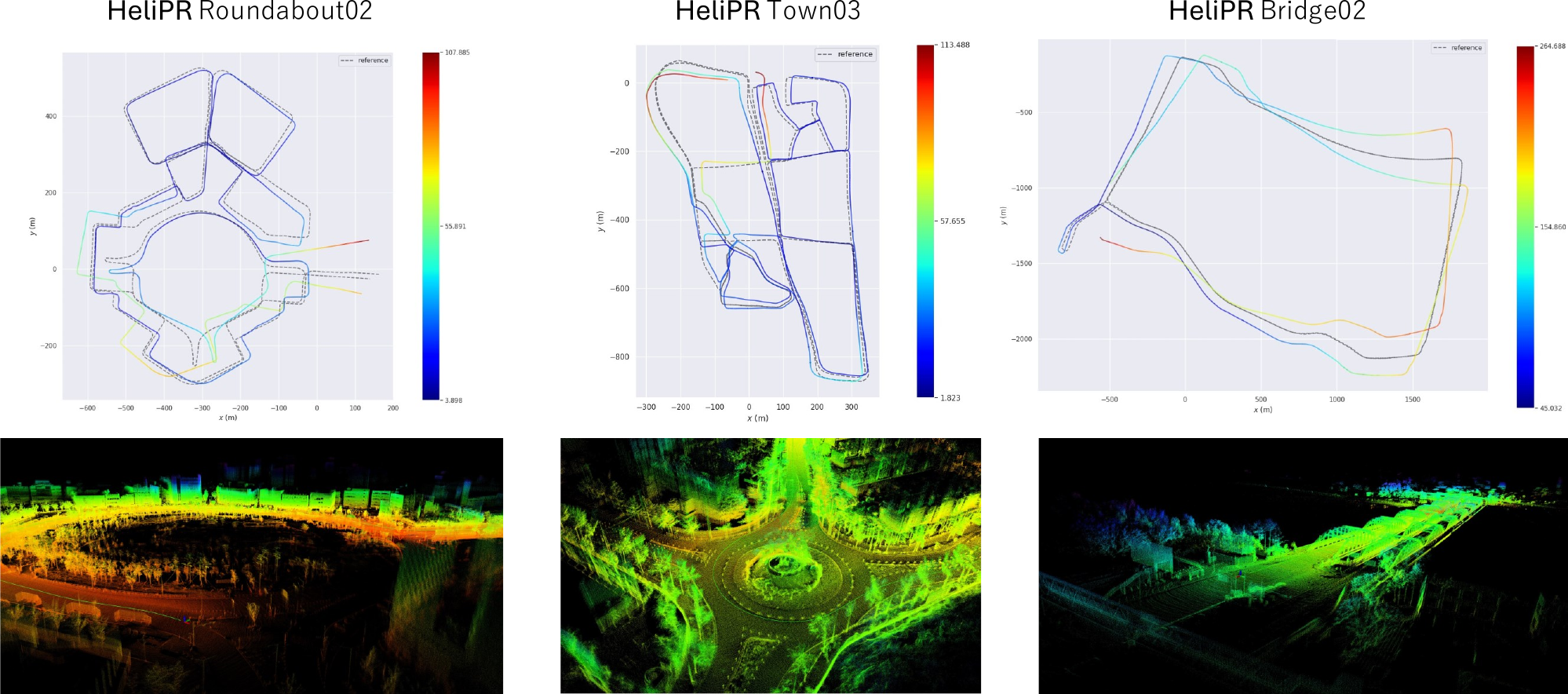}
    \caption{Paths and maps generated by the DualQuat-LOAM method with the ConSLAM, NTU VIRAL and HeLiPR datasets.} 
    \label{fig:figures_benchmark}
\end{figure*}

In the HeLiPR dataset, DualQuat-LOAM showed an error of $21.393 ~[m]$, a larger value compared to VoxelMap ($5.010 ~[m]$) and DLO ($5.022 m$). However, DualQuat-LOAM outperforms other methods such as KISS-ICP ($29.273 ~[m]$) and, although its error is high, it is still within an acceptable range for a LiDAR-only odometry method in a large urban environment. 

In addition to the experiments in Table 4, two additional tests were performed using the \texttt{Town03} and \texttt{Bridge02} sequences from the HeLiPR dataset, in order to evaluate the performance of the DualQuat-LOAM method under different conditions. The \texttt{Town03} sequence, recorded at night, presents a high complexity due to its dense urban environment at the Seochon location, with a path length of $8903~[m]$. In this scenario, DualQuat-LOAM obtained an ATE error of $19.791~[m]$. On the other hand, in the \texttt{Bridge02} sequence, captured during the daytime in and around Dongjak Bridge, with a path length of $14615~[m]$ and a large number of dynamic objects, DualQuat-LOAM showed an ATE error of $128.605~[m]$, indicating a significant degradation in performance due to the complexity of the scenario and the presence of moving vehicles. This experiments can be found in this video \footnote{\href{https://youtu.be/K1Jg5siXlD8?feature=shared}{https://youtu.be/K1Jg5siXlD8?feature=shared}}.

\begin{table}[h]
\footnotesize	
\centering
\caption{Benchmark results (up: LO, down: LIO). Evaluation Metric: ATE (m). The data extracted from the LiDAR odometry survey \cite{lee2024lidar}.}
\begin{tabular}{|l|c|c|c|}
\hline
\textbf{Algorithms} & \textbf{ConSLAM} & \textbf{NTU VIRAL} & \textbf{HeLiPR} \\ 
\hline
\textbf{LOAM} \cite{zhang2017loam} & - & 0.959 & 23.043 \\
\textbf{LeGO-LOAM} \cite{legoloam2018} & 0.263 & 8.478 & 10.111 \\
\textbf{KISS-ICP} \cite{vizzo2023kiss} & 13.517 & 0.829 & 29.273 \\
\textbf{DLO} \cite{DLO} & 0.154 & 0.142 & 5.022 \\
 \textbf{VoxelMap} \cite{yuan2022efficient} & - & 1.100 &5.010 \\
 \textbf{DualQuat-LOAM}& 5.334& 1.392&21.393\\
\hline
\textbf{LIO-SAM} \cite{LIO-SAM}& 0.167 & 0.115 & 42.414 \\
\textbf{FAST-LIO2} \cite{xu2022fast} & 0.113 & 0.116 & 1.655 \\
\textbf{Faster-LIO} \cite{bai2022faster} & 0.102 & 0.120 & 22.402 \\
\textbf{DLIO} \cite{chen2023direct} & 0.106 & 0.224 & 2.042 \\
\textbf{Point-LIO} \cite{he2023point} & 0.115 & 0.105 & 17.142 \\
\hline
\end{tabular}
\label{table:benchmark}
\end{table}

\section{Conclusions and Future Work}

We present the DualQuat-LOAM LiDAR odometry method, which is based on edge, surface and STD descriptor parameterization using dual quaternions. This approach allows to represent in a compact way the rotations and translations of a system, for this purpose it has been necessary also the parameterization of the optimizer in quaternions, thus ensuring a complete coherence in the pose estimation process. Results show that our method achieves pose estimation only with the point cloud of a LiDAR sensor, without the integration of additional sensors. Furthermore, our approach shows that the addition of a specific cost function for STD descriptors, each of which has its own reference frame with translation and rotation relative to a global reference frame, improves the performance of the pose estimation method compared to the exclusive use of edges and surfaces.

Currently, we have implemented the optimizer with Ceres using automatic differentiation, which generates limitations in the computational processing. Therefore, as future work, we plan to integrate the mathematical modeling necessary to compute the specific Jacobians of each cost function efficiently. In addition, we consider integrating our method with the fusion of other sensors, such as an IMU, using a Kalman filter parameterized in dual quaternions. These additions to the system could significantly aid in the accuracy, robustness and computational times of pose estimation, particularly in more complex and dynamic environments.

\section*{Acknowledgements}
This work has been supported by the grant PID2021-122685OB-I00 funded by MICIU/AEI/10.13039/501100011033 and by ERDF/EU, and grant PRE2019-088069 funded by MICIU/AEI/10.13039/501100011033 and ESF Investing in your future.

\bibliographystyle{cas-model2-names}
\bibliography{cas-refs}
\newpage
\bio{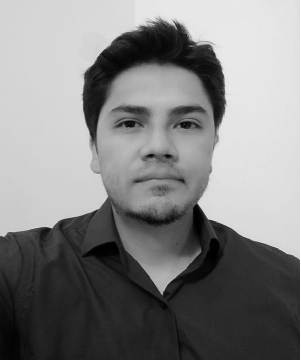}
\textbf{Edison P. Velasco Sánchez} received the degree in Electronic Engineering and Instrumentation from the University of the Armed Forces ESPE (Ecuador) in 2015 and a Master's Degree in Automation and Robotics from the University of Alicante (Spain) in 2018. He was a research technician the ESPE University and in the Automation, Robotics and Computer Vision Group (AUROVA) of the University of Alicante. He is currently pursuing a Ph.D. degree in AUROVA at the University of Alicante funding by the Regional Valencian Community Government and the Ministry of Science, Innovation and Universities  through the grant PRE2019-088069. His research interests include navigation and autonomous localization in UGVs with cameras and LIDAR sensors.
\endbio

\bio{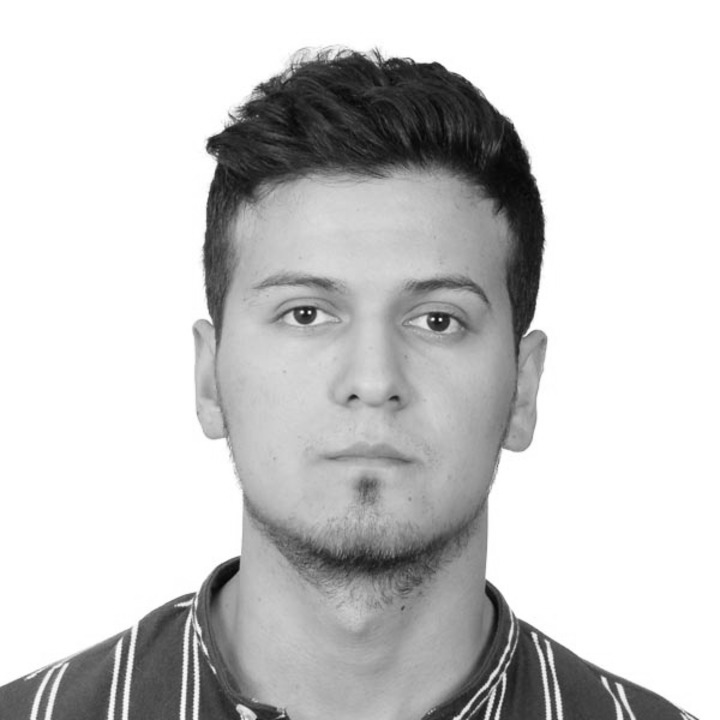}
\textbf{Luis F. Recalde} 
 graduated with a degree in Mechatronics Engineering from the University of the Armed Forces - ESPE in 2021 and is currently pursuing a master’s degree in Control Systems Engineering at the National University of San Juan, Argentina. His research centers on applying advanced control formulations, particularly Nonlinear Model Predictive Control (NMPC) and Quadratic Programming, to robotic systems. He has a strong interest in differential geometry for motion representation and control. In addition, he explores the integration of Reinforcement Learning (RL) to enhance control and planning algorithms in complex robotic applications.
\endbio

\bio{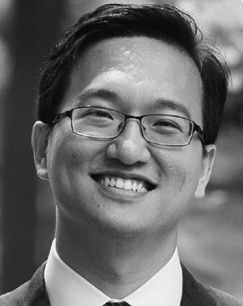}
\textbf{Guanrui Li} 
Guanrui Li (Graduate Student Member) received the bachelor's degree in theoretical and applied mechanics from Sun Yat-sen University, Guangzhou, China, where he was recognized as an undergraduate (Hons.), and the master's degree in robotics from the GRASP Lab, the University of Pennsylvania, Philadelphia, PA, USA. He received the Ph.D. degree in electrical and computer engineering with New York University, NY, USA, with a focus on robotics and aerial systems. He has an extensive publication record in top-tier robotics conferences and journals like ICRA, RA-L, and T-RO, and his work has garnered attention in various media, including IEEE Spectrum and the Discovery Channel. His research interests include the dynamics, planning, and control of robotics systems, with applications in aerial transportation and manipulation, as well as human–robot collaboration. Dr. Guanrui was the recipient of several notable recognitions, including the NSF CPS Rising Stars in 2023, the Outstanding Deployed System Paper Award finalist at 2022 IEEE ICRA, and the 2022 Dante Youla Award for Graduate Research Excellence at NYU.
\endbio

\bio{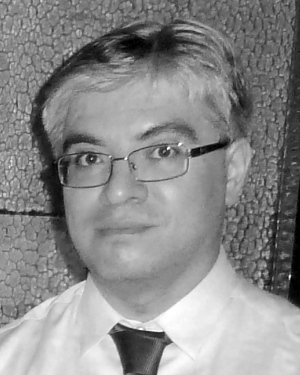}
\textbf{Francisco A. Candelas} received the Computer Science Engineer and the Ph.D. degrees in the University of Alicante (Spain), in 1996 and 2001 respectively. He is Associate Professor in the University of Alicante since 2003, where he teaches currently courses about Automation and Robotics Sensors in the Degree in Robotic Engineering. Previously, he was in tenure track from 1999 to 2003. Dr. Candelas also researches in the Automation, Robotics and Computer Vision Group (AUROVA) of the University of Alicante since 1998, and he has involved in several research projects and networks supported by the Spanish Government, as well as development projects in collaboration with regional industry. His main research topics are autonomous robots, robot development, and virtual/remote laboratories for teaching.
\\
\endbio

\bio{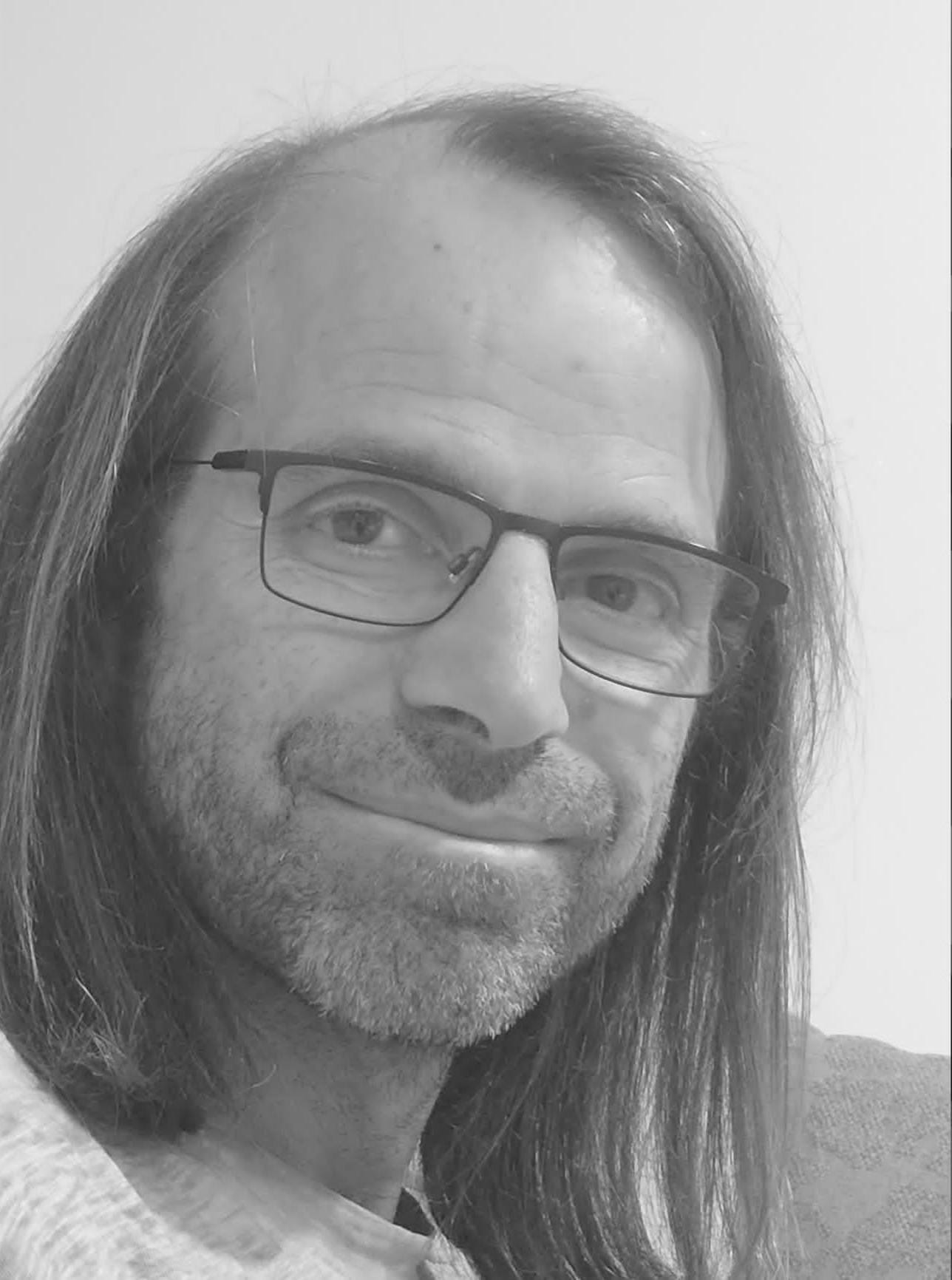}\textbf{Santiago T. Puente} received the Computer Science Engineer and the PhD degrees in the University of Alicante (Spain), in 1998 and 2003 respectively. He is a full-time lecture and researcher at the University of Alicante since 2003, where he teaches currently in the Degree in Robotic Engineering. From 2013 to 2019, Dr. Puente has been deputy director of Infrastructures and facilities Polytechnic School of the University of Alicante, Furthermore, from 2019 to 2021 he has been Academic Coordinator  of BEng Robotics Engineering. Dr. Puente also researches in the Automation, Robotics and Computer Vision Group (AUROVA) of the University of Alicante since 1999, and he has involved in several research projects and networks supported by the Spanish Government, as well as development projects in collaboration with regional industry. His research interests include automation and robotics (intelligent robotic manipulation, robot perception systems, robot imitation learning, field mobile robots), and e-learning. Currently, his research focuses on robot imitation learning, and e-learning.
\endbio

\bio{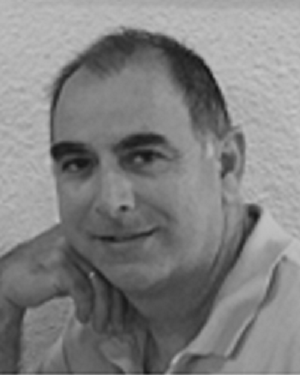} 
\textbf{Fernando Torres} was born in Granada, where he attended primary and high school. He moved to Madrid to undertake a degree in Industrial Engineering at the Polytechnic University of Madrid, where he also carried out his PhD thesis. The last year of his PhD thesis he became a full-time lecturer and researcher at the University of Alicante, and he has worked there ever since. He directs the research group ``Automatics, Robotics and Computer Vision'' founded in 1996 at the University of Alicante. He is a member of TC 5.1 and TC 9.4 of the IFAC, a Senior Member of the IEEE and a member of CEA. Since july 2018 he is coordinator of the area of Electrical, Electronic and Automatic (IEA) of the Spanish Agency of Statal Research (AEI). His research interests include automation and robotics (intelligent robotic manipulation, visual control of robots, robot perception systems, field mobile robots, advanced automation for industry 4.0, artificial vision engineering), and e-learning. Currently, his research focuses on automation, robotics, and e-learning. In these lines, it currently has more than fifty publications in JCR-ISI journals and more than a hundred papers in international congresses. He was Leader Research in several research projects and he has supervised several PhD in these lines of research.
\endbio

\end{document}